\documentclass[lettersize,journal]{IEEEtran}

\AtBeginDocument{%
  \providecommand\BibTeX{{%
    \normalfont B\kern-0.5em{\scshape i\kern-0.25em b}\kern-0.8em\TeX}}}
    
\newif\ifarxiv
\arxivfalse

\usepackage{amsmath,amsfonts}
\usepackage{algorithmic}
\usepackage{algorithm}
\usepackage{array}
\usepackage{textcomp}
\usepackage{stfloats}
\usepackage{url}
\usepackage{verbatim}
\usepackage{graphicx}

\usepackage{subfigure}
\usepackage{subcaption}
\usepackage{cite}
\usepackage{hyperref}
\usepackage[all]{hypcap}
\hypersetup{
	colorlinks=true,
	linkcolor=black,
	filecolor=blue,      
	urlcolor=black,
	citecolor=black,
}
\usepackage{xcolor}
\usepackage{xspace}
\usepackage{colortbl}
\usepackage{hhline}
\usepackage{ulem}
\usepackage{soul}

\usepackage{booktabs}
\usepackage{adjustbox}
\usepackage{makecell}
\usepackage{tabularx}
\usepackage{xcolor}
\definecolor{c4}{RGB}{255,225,187}
\definecolor{c2}{RGB}{209, 233, 184}
\definecolor{c3}{RGB}{218,243,246}
\definecolor{c1}{RGB}{249, 229, 229}
\definecolor{c5}{RGB}{255, 128, 128}
\definecolor{c6}{RGB}{242, 255, 128}
\usepackage{tikz}
\usetikzlibrary{shadings}
\usetikzlibrary{shapes.geometric, backgrounds}

\newcommand{\customstrut}[2]{\vrule height #1 depth #2 width 0pt}
\newcommand{\GradientText}[1]{
    \begin{tikzpicture}[baseline] 
        \node[inner sep=\fboxsep, anchor=base, left color=c1, right color=c2]{\customstrut{3pt}{1pt} #1};
    \end{tikzpicture}
}
\newcommand{\GradientTwo}[1]{
    \begin{tikzpicture}[baseline] 
        \node[inner sep=\fboxsep, anchor=base, left color=c2, right color=c3] {\customstrut{3pt}{1pt} #1};
    \end{tikzpicture}
}

\newcommand{\one}{$^\dagger$}  
\newcommand{\two}{$^\ddagger$}  
\newcommand{\three}{$^\mathsection$} 
\newcommand{\four}{$^\|$} 
\newcommand{\five}{$^\mathparagraph$} 

\usepackage{makecell}
\usepackage{adjustbox} 
\usepackage{lipsum} 
\renewcommand{\arraystretch}{1.5}
\usepackage{etoolbox}

\begin{document}

\title{Retrieval-Augmented Generation for\\AI-Generated Content: A Survey}
\author{Penghao Zhao{$^{*}$}, Hailin Zhang{$^{*}$}, Qinhan Yu, Zhengren Wang, Yunteng Geng, \\ Fangcheng Fu{$^\dagger$}, Ling Yang, Wentao Zhang{$^\dagger$}, Jie Jiang, Bin Cui{$^\dagger$}
\thanks{$^{*}$ Both authors contributed equally to this research.}
\thanks{$^\dagger$ Corresponding authors.}
\thanks{$\bullet$ Penghao Zhao, Hailin Zhang, Qinhan Yu, Zhengren Wang, Yunteng Geng, Fangcheng Fu, Ling Yang, Wentao Zhang and Bin Cui are with Peking University (e-mail: penghao.zhao@stu.pku.edu.cn, z.hl@pku.edu.cn, yuqinhan@stu.pku.edu.cn, wzr@stu.pku.edu.cn, 1800012997@pku.edu.cn, ccchengff@pku.edu.cn, yangling0818@163.com, wentao.zhang@pku.edu.cn,  bin.cui@pku.edu.cn).}
\thanks{$\bullet$ Jie Jiang is with Tencent Inc. (email: zeus@tencent.com)}
}

\maketitle

\begin{abstract}
Advancements in model algorithms, the growth of foundational models, and access to high-quality datasets have propelled the evolution of Artificial Intelligence Generated Content (AIGC).
Despite its notable successes, AIGC still faces hurdles such as updating knowledge, handling long-tail data, mitigating data leakage, and managing high training and inference costs.
Retrieval-Augmented Generation (RAG) has recently emerged as a paradigm to address such challenges.
In particular, RAG introduces the information retrieval process, which enhances the generation process by retrieving relevant objects from available data stores, leading to higher accuracy and better robustness.
In this paper, we comprehensively review existing efforts that integrate RAG techniques into AIGC scenarios.
We first classify RAG foundations according to how the retriever augments the generator,
distilling the fundamental abstractions of the augmentation methodologies for various retrievers and generators.
This unified perspective encompasses all RAG scenarios, illuminating advancements and pivotal technologies that help with potential future progress.
We also summarize additional enhancements methods for RAG, facilitating effective engineering and implementation of RAG systems.
Then from another view, we survey on practical applications of RAG across different modalities and tasks, offering valuable references for researchers and practitioners.
Furthermore, we introduce the benchmarks for RAG, discuss the limitations of current RAG systems, and suggest potential directions for future research.
Github: \href{https://github.com/PKU-DAIR/RAG-Survey}{{https://github.com/PKU-DAIR/RAG-Survey}}.

\end{abstract}
\vspace{-0.1cm}


\section{\textbf{Introduction}}
\label{Intro}

\ifarxiv
\begin{figure*}[t]
\centering
\includegraphics[width=0.8\textwidth]{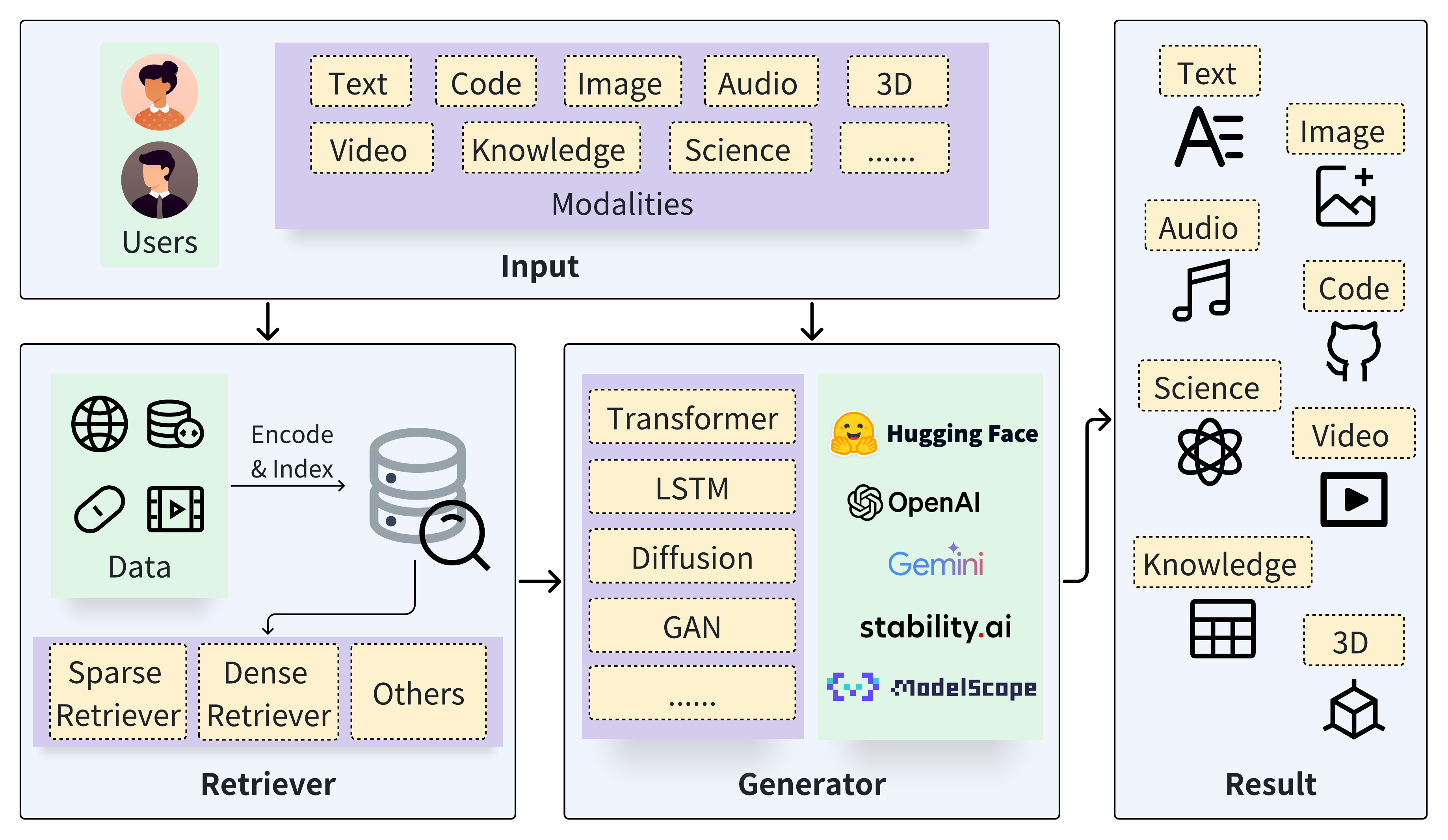}
\caption{A generic RAG architecture. The user queries, spanning different modalities, serve as input to both the retriever and the generator. The retriever extracts relevant information from data sources. The generator interacts with the retrieval results and ultimately produces outcomes of various modalities.}
\label{fig:overview}
\end{figure*}

\else
\begin{figure*}[t]
\centering
\includegraphics[width=0.8\linewidth]{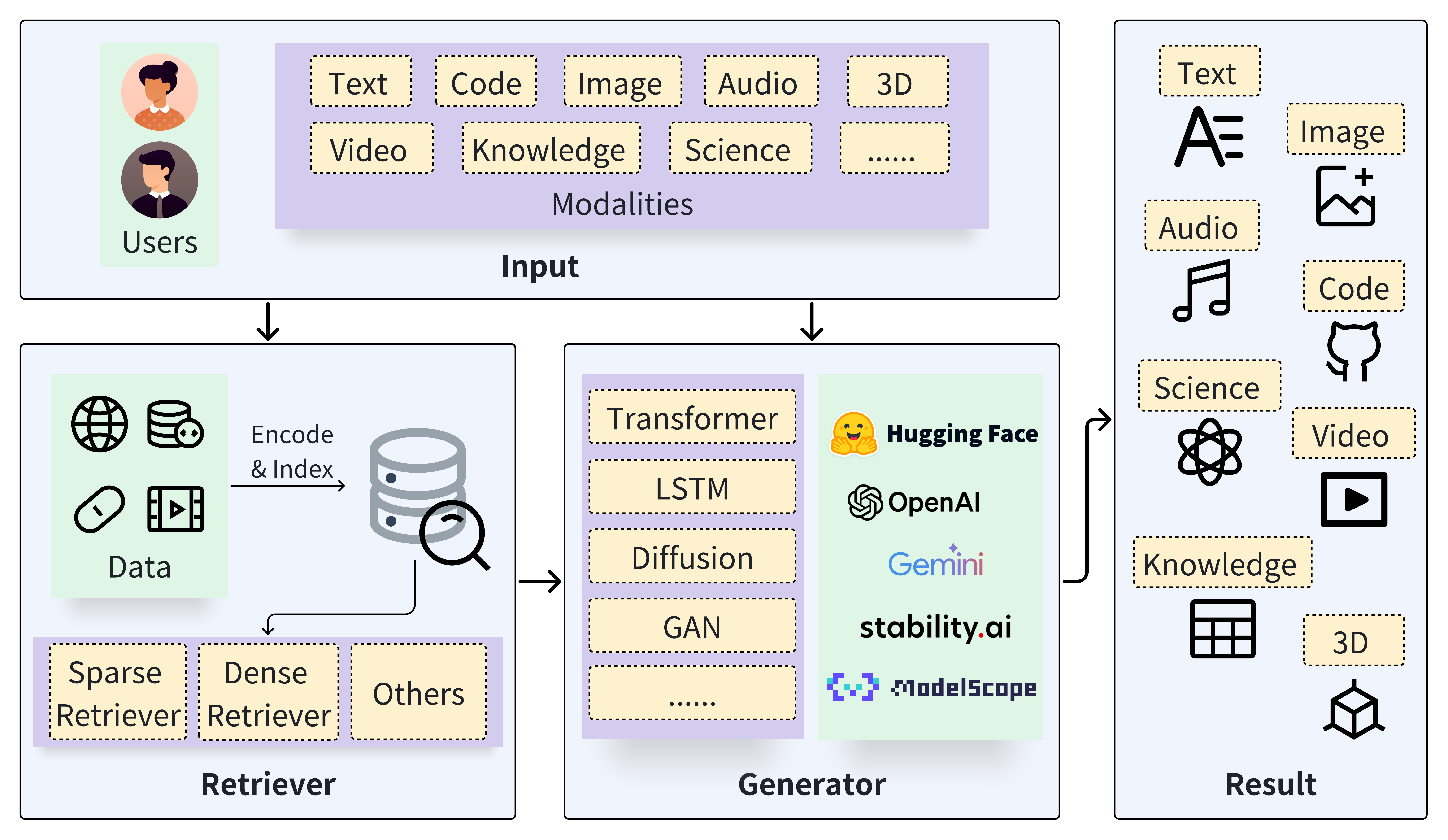}
\caption{A generic RAG architecture. The user queries, spanning different modalities, serve as input to both the retriever and the generator. The retriever extracts relevant information from data sources. The generator interacts with the retrieval results and ultimately produces outcomes of various modalities.}
\label{fig:overview}
\end{figure*}
\fi

\subsection{\textbf{Background}}

Recent years have witnessed the surge in interests surrounding Artificial Intelligence Generated Content (AIGC). 
Various content generation tools have been meticulously crafted to produce diverse outputs across various modalities, such as Large Language Models (LLMs) including the GPT series~\cite{DBLP:conf/nips/BrownMRSKDNSSAA20,DBLP:journals/corr/abs-2107-03374,DBLP:journals/corr/abs-2303-08774} and the LLAMA series~\cite{LLaMA,DBLP:journals/corr/abs-2307-09288,DBLP:journals/corr/abs-2308-12950} for texts and codes, DALL-E~\cite{DBLP:conf/icml/RameshPGGVRCS21,DBLP:journals/corr/abs-2204-06125,betker2023improving} and Stable Diffusion~\cite{DBLP:conf/cvpr/RombachBLEO22} for images, and Sora~\cite{openai/sora} for videos.
The word ``AIGC'' emphasizes that the contents are produced by advanced generative models other than human beings or rule-based approaches.
These generative models have achieved remarkable performance due to the utilization of novel model algorithms, explosive scale of foundation models, and massive high-quality datasets.
Specifically, sequence-to-sequence tasks have transitioned from utilizing Long Short-Term Memory (LSTM) networks~\cite{DBLP:journals/neco/HochreiterS97} to Transformer-based models~\cite{DBLP:conf/nips/VaswaniSPUJGKP17}, and image-generation tasks have shifted from Generative Adversarial Networks (GANs)~\cite{GAN} to Latent Diffusion Models (LDMs)~\cite{DBLP:conf/cvpr/RombachBLEO22} as well.
Notably, the architecture of foundation models, initially constituted by millions of parameters~\cite{DBLP:conf/naacl/DevlinCLT19,DBLP:journals/jmlr/RaffelSRLNMZLL20}, has now grown to billions or even trillions of parameters~\cite{LLaMA,DBLP:conf/nips/BrownMRSKDNSSAA20,Switch_transformers}.
These advancements are further bolstered by the availability of rich, high-quality datasets~\cite{DBLP:conf/nips/BrownMRSKDNSSAA20,scalingLaw}, which provide ample training samples to fully optimize model parameters.

Information retrieval is another pivotal application within the field of computer science.
Different from generation, retrieval aims to locate relevant existing objects from a vast pool of resources.
The most prevalent application of retrieval lies in web search engines, which primarily focus on the task of document retrieval~\cite{DBLP:journals/ftir/RobertsonZ09,DBLP:conf/emnlp/KarpukhinOMLWEC20}. 
In the present era, efficient information retrieval systems can handle document collections on the order of billions~\cite{DBLP:journals/tbd/JohnsonDJ21,DBLP:conf/nips/ChenZWLLLYW21}.
Besides documents, retrieval has also been applied for many other modalities~\cite{DBLP:journals/csur/DattaJLW08,radford2021learning,DBLP:conf/emnlp/FengGTDFGS0LJZ20,DBLP:conf/icassp/WuCZHBD23}.

Despite significant advancements in generative models, AIGC still grapples with challenges like outdated knowledge, lack of long-tail knowledge~\cite{DBLP:conf/acl/MallenAZDKH23}, and risks of leaking private training data~\cite{DBLP:conf/uss/CarliniTWJHLRBS21}.
Retrieval-Augmented Generation (RAG) aims to mitigate these issues with its flexible data repository~\cite{C-RAG}. The retrievable knowledge acts as non-parametric memory, which is easily updatable, accommodates extensive long-tail knowledge, and can encode confidential data.
Moreover, retrieval can lower generation costs. RAG can reduce the size of large models~\cite{Atlas}, support long contexts~\cite{MemTransformer2022}, and eliminate certain generation steps~\cite{REST}.

A typical RAG process is depicted in Fig.~\ref{fig:overview}. 
Given an input query, the retriever identifies relevant data sources, and the retrieved information interacts with the generator to improve the generation process.
%
There are several \ul{\textit{foundational paradigms}} (\ul{\textit{foundations}} in short) according to how the retrieved results augment the generation:
they can serve as augmented input to the generator~\cite{REALM,2020RAG}; they can join at the middle stage of generation as latent representations~\cite{FId,RETRO}; they can contribute to the final generation results in the form of logits~\cite{KNN-LM,Efficient-KNNLM}; they can even influence or omit certain generation steps~\cite{REST,GPTCache}.
%
Additionally, researchers have proposed various \ul{\textit{enhancements}} to improve the foundational RAG process.
These methods encompass specific optimizations for individual components as well as holistic enhancements aimed at the entire pipeline.

In addition, while the concept of RAG initially emerged in text-to-text generation~\cite{2020RAG}, this technique has also found \ul{\textit{applications}} across various domains, including codes~\cite{DBLP:conf/emnlp/ParvezACRC21,DBLP:conf/naacl/AhmadCRC21,DBLP:conf/iclr/Zhou0XJN23}, audios~\cite{DBLP:journals/corr/abs-2012-07331,DBLP:conf/icml/HuangHY0LLYLYZ23}, images~\cite{tseng2020retrievegan,sarto2022retrieval,ramos2023smallcap}, videos~\cite{DBLP:journals/tomccap/ChenPLYCM23,DBLP:journals/corr/abs-2401-00789}, 3D~\cite{DBLP:journals/corr/abs-2402-02972,DBLP:conf/iccv/ZhangGPCHLYL23}, knowledge~\cite{DBLP:conf/coling/HuWSQ22,DBLP:conf/emnlp/HuangKZ21,DBLP:conf/emnlp/DasZTGPLTPM21}, and AI for science~\cite{wang2022retrieval,jin2023genegpt}.
In particular, the essential idea and process of RAG are largely consistent across modalities.
However, it necessitates minor adjustments in augmentation techniques, and the selection of retrievers and generators varies depending on the specific modalities and applications.

Despite the rapid growth in recent research on RAG and the booming applications, a systematic review encompassing all foundations, enhancements, and applications is notably absent, hindering the development of this field.
For one thing, the absence of discussion on RAG foundations significantly undermines the practical value of the research in this domain, leaving the potential of RAG not fully explored.
While the majority of research interest, particularly among LLM researchers, centers on query-based RAG in text-generation tasks, it is essential to acknowledge that other RAG foundations are also effective and with significant potential for usage and further development.
For another, the lack of an overview on RAG applications causes researchers and practitioners to overlook RAG's progress across multiple modalities and remain unaware of how RAG can be effectively applied.
Although text generation is typically considered as the main application of RAG, we emphasize that the development of RAG in other modalities has also begun to catch on and has yielded promising advancements.
Certain modalities have a rich historical connection to retrieval techniques, infusing RAG with distinctive characteristics.
Inspired by this, in this paper, our objective is to present a comprehensive survey to provide a systematic overview of RAG.
\subsection{\textbf{Contribution}}
This survey offers a comprehensive overview of RAG, covering foundations, enhancements, applications, benchmarks, limitations, and potential future directions.
Despite variations in retrievers and generators across modalities and tasks, we distill the core principles of RAG foundations, viewing applications as adaptations of these principles.
We aim to offer references and guidelines to researchers and practitioners, providing valuable insights for advancing RAG methodologies and related applications.
In summary, we list our contributions as follows:
\begin{itemize}
    \item We conduct a comprehensive review of RAG, and distill the abstractions of RAG foundations for various retrievers and generators.
    \item We investigate the enhancements in the literature of RAG, elaborating the techniques leveraged to enable more effective RAG systems.
    \item For various modalities and tasks, we survey existing AIGC methods that incorporate RAG techniques, exhibiting how RAG contributes to current generative models.
    \item We discuss the limitations and promising research directions of RAG, shedding light on its potential future development.
\end{itemize}
\subsection{\textbf{Related Work}}

As the field of RAG advances, several surveys have emerged; yet they address only specific facets of the area. In particular, they either exclusively focus on a single RAG foundation or provide only a brief overview of RAG augmentation methodologies for limited scenarios.

Most of the existing works focus on text-related RAG tasks that are facilitated by LLMs, without in-depth investigation in other modalities.
The survey by Li et al.~\cite{DBLP:journals/corr/abs-2202-01110} offers a basic overview of RAG and discusses specific applications within the scope of text generation tasks.
In a similar vein, the tutorial crafted by Asai et al.~\cite{retrieval-lm-tutorial} centers on retrieval-based language models, detailing their structures and training strategies.
Meanwhile, a recent survey by Gao et al.~\cite{DBLP:journals/corr/abs-2312-10997} explores RAG in the context of LLMs, with a particular emphasis on enhancement approaches for query-based RAG.
Recognizing that RAG has extended beyond the text domain, our work broadens its reach to the entire AIGC landscape, facilitating a more comprehensive coverage of RAG research.

In addition, another survey proposed by Zhao et al.~\cite{DBLP:conf/emnlp/ZhaoCWJLQDGLLJ23} introduces RAG applications across multiple modalities, but ignoring the discussion on RAG foundations.
Another work~\cite{ding2024survey} covers only part works of other modalities.
While existing research has explored various aspects of RAG, there remains a need for a comprehensive overview that covers RAG foundations, enhancements, and its applicability across different domains.
In this paper, we aim to address the gap by presenting a systematic survey of RAG.

\subsection{\textbf{Roadmap}}

The rest of the paper is organized as follows. 
Section~\ref{sec:preliminary} elaborates on the preliminary of RAG, introducing retrievers and generators. 
Section~\ref{sec:methods} presents RAG foundations and further enhancements on RAG. 
Section~\ref{sec:applications} reviews existing research on RAG across various applications. 
Section~\ref{sec:benchmark} investigates the benchmark frameworks for RAG.
Section~\ref{sec:discussion} discusses current limitations of RAG and potential future directions. 
Finally, Section~\ref{sec:conclusion} concludes this paper.

\section{\textbf{Preliminary}}
\label{sec:preliminary}

In this section, we provide an overview of the general RAG architecture and explore the generators and the retrievers in today’s RAG-based AIGC.
\subsection{\textbf{Overview}}
\label{pre_overview}

As shown in Fig.~\ref{fig:overview}, the entire RAG system consists of two core modules: the retriever and the generator,
where the retriever searches for relevant information from the data store and the generator produces the required contents.
The RAG process unfolds as follows: (i) the retriever initially receives the input query and searches for relevant information; (ii) then, the original query and the retrieval results are fed into the generator through a specific augmentation methodology; (iii) finally, the generator produces the desired outcomes.
\subsection{\textbf{Generator}}
\label{pre_gen}
\begin{figure*}[!t]
  \centering
  \includegraphics[width=0.9\linewidth]{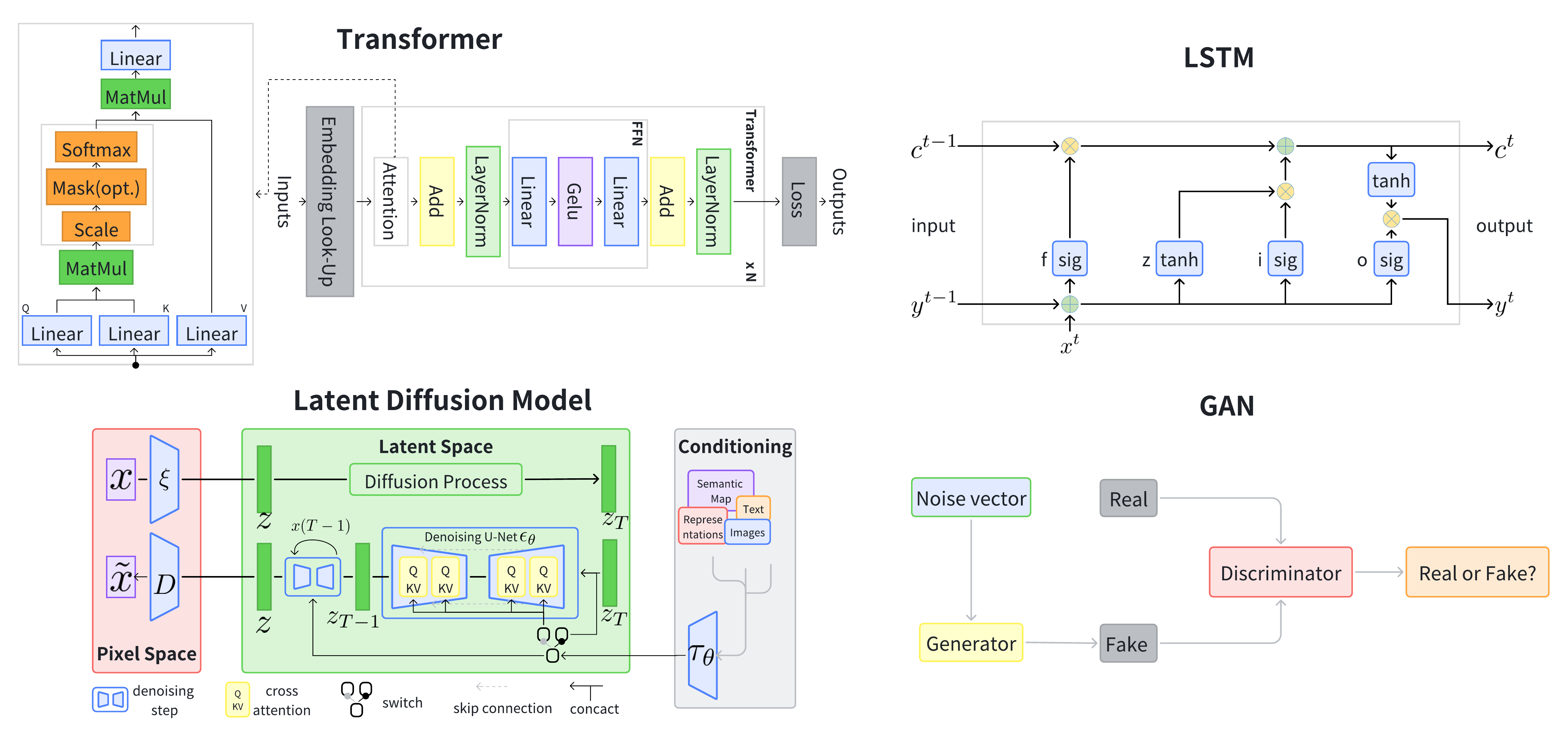}
  \caption{General architectures of several generators.}
  \label{fig:generators}

\end{figure*}
The remarkable performance of generative AI across diverse tasks has ushered in the era of AIGC.
The generation module plays a crucial role within the RAG system.
Different generative models are applied for different scenarios, such as transformer models for text-to-text tasks, VisualGPT~\cite{DBLP:conf/cvpr/ChenGY0E22} for image-to-text tasks, Stable Diffusion~\cite{DBLP:conf/cvpr/RombachBLEO22} for text-to-image tasks, Codex~\cite{DBLP:journals/corr/abs-2107-03374} for text-to-code tasks, etc.
Here we introduce 4 typical generators that are frequently used in RAG: transformer model, LSTM, diffusion model, and GAN. 

\subsubsection{\textbf{Transformer Model}}
Transformer models are one of the best performing models in the field of Natural Language Processing~(NLP), consisting of self-attention mechanisms, feed-forward networks, layer normalization modules, and residual networks~\cite{EfficientTransformers}.
As illustrated in Figure~\ref{fig:generators}, the final output sequence is produced by executing vocabulary classification at each generative step on a sequence of latent representations derived from tokenization and embedding.

\subsubsection{\textbf{LSTM}}

As shown in Fig.~\ref{fig:generators}, Long Short-Term Memory (LSTM)~\cite{lstm_survey} is a special form of Recurrent Neural Network (RNN) model.
It tackles the issues of exploding/vanishing gradients in long-term dependency processing by incorporating cell states and gating mechanisms.
The model comprises three gates (Input, Forget, and Output) that filter information, and a central Cell State module that retains and manages information.
It uses the same vocabulary classification method as transformer models to autoregressively generate outputs.

\subsubsection{\textbf{Diffusion Model}}

Diffusion models are a family of deep generative models that can create realistic and diverse samples of data (including images, texts, videos, molecules, etc.)~\cite{yang2023diffsurvey}.
As shown in Fig.~\ref{fig:generators}, diffusion models work by gradually adding noise to data until it becomes random, then reversing the process to generate new data from noise. 
This process is based on probabilistic modeling and neural networks.
\subsubsection{\textbf{GAN}}

Generative Adversarial Networks (GANs)~\cite{GAN} are highly anticipated deep learning models which can simulate and generate realistic images, audio, and other data~\cite{GAN_Survey}. 
As shown in Fig.~\ref{fig:generators}, a typical GAN consists of two main components: a generator and a discriminator.
These two parts compete with each other through adversarial learning, allowing the generator to continuously improve its ability to generate realistic samples, while the discriminator continuously improves its ability to distinguish between true and false samples.

\subsection{\textbf{Retriever}}
\label{pre:retrieve}
Retrieval is to identify and obtain relevant information given an information need. 
Specifically, let’s consider information resources that can be conceptualized as a key-value store, where each key corresponds to a value (keys and values can be identical). 
Given a query, the objective is to search for the top-$k$ most similar keys using a similarity function, and obtain the paired values.
Based on different similarity functions, existing retrieval methods can be categorized into sparse retrieval, dense retrieval, and others.
In widely used sparse and dense retrieval, the entire process can be divided into two distinct phases: (i) each object is first encoded into a specific representation; and then (ii) an index is constructed to organize the data source for efficient search.

\subsubsection{\textbf{Sparse Retriever}}

Sparse retrieval methods are commonly used in document retrieval, where the keys/values represent the documents to be searched. 
These methods leverage term matching metrics such as TF-IDF~\cite{DBLP:conf/sigir/RobertsonW97}, query likelihood~\cite{DBLP:conf/sigir/LaffertyZ01}, and BM25~\cite{DBLP:journals/ftir/RobertsonZ09}, which analyze word statistics from texts and construct inverted indices for efficient searching.
Essentially, BM25 is a strong baseline in large-scale web search, integrating inverse document frequency weights, query token occurrences, and other pertinent metrics.

To enable efficient search, sparse retrieval typically leverages an inverted index to organize documents.
Concretely, each term from the query performs a lookup to obtain a list of candidate documents, which are subsequently ranked based on their statistical scores.

\subsubsection{\textbf{Dense Retriever}}

Unlike sparse retrieval, dense retrieval methods represent queries and keys using dense embedding vectors, and build Approximate Nearest Neighbor (ANN) index to speed up the search.
This can be applied to all modalities.
For text data, recent advancements in pre-trained models (such as BERT~\cite{DBLP:conf/naacl/DevlinCLT19}) have been employed encode queries and keys individually~\cite{DBLP:conf/emnlp/KarpukhinOMLWEC20}. This approach is often referred to as Dense Passage Retrieval (DPR).
Similar to text, models have been proposed to encode code data~\cite{DBLP:conf/emnlp/FengGTDFGS0LJZ20}, audio data~\cite{DBLP:conf/icassp/HersheyCEGJMPPS17}, image data~\cite{radford2021learning}, video data~\cite{DBLP:conf/cvpr/DongLXJH0W19}, etc.
The similarity score between dense representations are usually computed with metrics such as cosine, inner product, L2-distance.

During training, dense retrieval uses contrastive learning to increase the similarity of positive samples and decrease that of negative ones. Several hard negative techniques~\cite{DBLP:conf/iclr/XiongXLTLBAO21} have been proposed to further enhance model quality. For efficient searching during inference, ANN methods are employed. Various indices are developed to serve ANN search, including tree~\cite{bentley1975multidimensional,li2023learning}, locality sensitive hashing~\cite{datar2004locality}, neighbor graph indices (e.g., HNSW~\cite{malkov2018efficient}, DiskANN~\cite{jayaram2019diskann}), and combined graph and inverted indices (e.g., SPANN~\cite{DBLP:conf/nips/ChenZWLLLYW21}).

\subsubsection{\textbf{Others}}
In addition to sparse retrieval and dense retrieval, there are alternative methods for retrieving relevant objects~\cite{DBLP:conf/nips/WangHWMWCXCZL0022,DBLP:conf/nips/ZhangWCCZMHDMWP23}.
Instead of calculating representations, some research works directly use the edit distance between natural language texts~\cite{DBLP:conf/emnlp/HayatiOAYTN18} or abstract syntax trees (AST) of code snippets~\cite{DBLP:conf/icse/ZhangW00020,DBLP:conf/iclr/PoesiaP00SMG22}.
In knowledge graphs, entities are connected by relations, serving as a pre-built index for retrieval. Thus, RAG methods utilizing knowledge graphs can employ $k$-hop neighbor searches for retrieval~\cite{DBLP:conf/acl/YeYHZX22,DBLP:journals/corr/abs-2210-12925}. Another retrieval method is Named Entity Recognition (NER)~\cite{lin2020bridging}, where the query is the input and the entities act as keys.

\section{\textbf{Methodologies}}
\label{sec:methods}

In this section, we first introduce foundational paradigms of RAG, and then outline enhancement methods that further improve the effectiveness.
\subsection{\textbf{RAG Foundations}}
\label{sec:methodologies}

\begin{figure*}
\centering
\includegraphics[width=1.0 \textwidth]{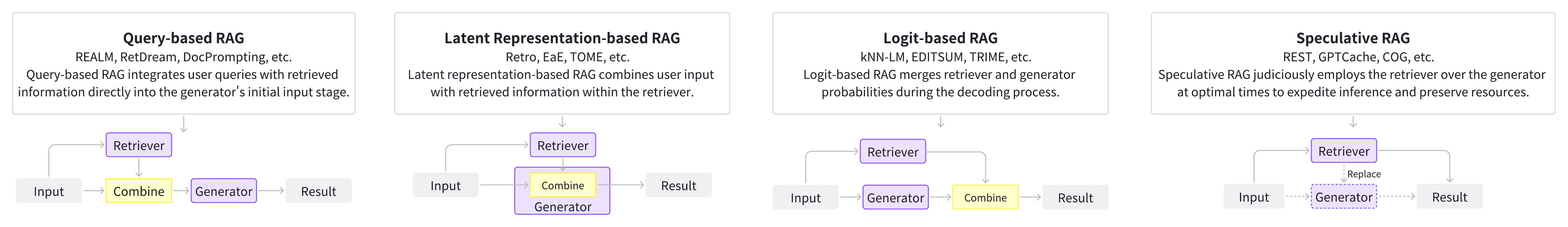}
\caption{Taxonomy of RAG foundations.}
\label{fig:Augmentation}
\end{figure*}

Based on how the retriever augments the generator, we categorize RAG foundations into 4 classes, as shown in Fig.~\ref{fig:Augmentation}.

\subsubsection{\textbf{Query-based RAG}}
Stemming from the idea of prompt augmentation, query-based RAG seamlessly integrates the user's query with insights from retrieved information, feeding it directly into the initial stage of the generator's input. 
This method is prevalent in RAG applications. Post-retrieval, the obtained content is merged with the user's original query to form a composite input, which is then processed by the generator to create a response. 
Query-based RAG is widely employed across various modalities.

For text generation, REALM~\cite{REALM} employs a dual-BERT framework to streamline knowledge retrieval and integration, marrying pre-trained models with knowledge extractors.
Lewis et al.~\cite{2020RAG} leveraged DPR for information retrieval and employs BART as the generator 
to effectively enhance the generation. 
SELF-RAG~\cite{Self-RAG} utilizes a critique module to determine whether the retrieval is required.
In addition to being compatible with local generators, query-based RAG is also applicable to scenarios that use LLM through API calls.
REPLUG~\cite{REPLUG} follows this methodology by treating the language model as a ``black box'', and effectively integrates relevant external documents into the query.
In-Context RALM~\cite{RALM} uses BM25 for document retrieval and trains a predictive reranker  to reorder and integrate the top-ranked documents.

In the field of code, several works~\cite{DBLP:conf/emnlp/ZanCLGWL22,DBLP:conf/iclr/Zhou0XJN23,DBLP:conf/icse/NashidSM23,DBLP:conf/sigsoft/JinSTSLSS23,DBLP:conf/acl/LuDHGHS22} have utilized the query-based paradigm to incorporate contextual information from text or code into the prompt, resulting in improved effectiveness of downstream tasks.

Recent researches in Knowledge Base Question Answering (KBQA) has also shown significant effects of combining retrieval and language models. For instance, Uni-Parser~\cite{DBLP:conf/emnlp/LiuYMRXZ22}, RNG-KBQA~\cite{DBLP:conf/acl/YeYHZX22}, and ECBRF~\cite{DBLP:conf/eacl/YangDCC23} effectively improve the performance and accuracy of QA systems by merging queries and retrieved information into prompts. 

In the AI-for-Science field, Chat-Orthopedist~\cite{shi2023retrieval} aids shared decision-making for adolescents with idiopathic scoliosis, improving LLMs' effectiveness and information precision by incorporating retrieved data into model prompts.

In the image generation task, RetrieveGAN~\cite{tseng2020retrievegan} boosts the relevance and precision of generated images by incorporating retrieved data, such as selected image patches and their bounding boxes, into the generator's input stage.
IC-GAN~\cite{casanova2021instance} modulates the specific conditions and details of the generated images by concatenating noise vectors with instance features.

For 3D generation, RetDream~\cite{DBLP:journals/corr/abs-2402-02972} initially utilizes CLIP~\cite{radford2021learning} to retrieve relevant 3D assets, then merges the retrieved contents with the user input during the input phase. 

Query-based RAG, often paired with LLM generators, offers modular flexibility, allowing swift integration of pre-trained components for quick deployment. Prompt design is crucial for utilizing retrieved data within this setup.
\subsubsection{\textbf{Latent Representation-based RAG}}

In latent representation-based RAG framework, retrieved objects are incorporated into generative models as latent representations. This enhances the model’s comprehension abilities and improves the quality of the generated content.

In the text field, FiD~\cite{FId} and RETRO~\cite{RETRO} are two classic structures of latent representation-based RAG, with many subsequent works conducting modifications based on them. 
FiD~\cite{FId} processes each retrieved paragraph and its title alongside the query through distinct encoders, then amalgamates the resulting latent representations for decoding by a single decoder to produce the final output.
RETRO~\cite{RETRO} retrieves relevant information for each segmented sub-query, then applies a novel module termed Chunked Cross-Attention (CCA) to integrate the retrieved contents with each sub-query tokens. 
In addition, there are other noteworthy novel structures within the scope of latent representation-based RAG.
Several studies~\cite{MemTransformer2022,Unlimiformer} have integrated k Nearest Neighbor (kNN) search within transformer blocks, allowing for input chunking and, in theory, addressing the long-criticized context length constraints of Transformer models.
Kuratov et al.~\cite{RMT-R} integrated Transformer with RNN, utilizing the model's intermediate output as the content for retrieval.

In the realms of code and science, FiD has gained widespread adoption, with applications spanning various code-related fields~\cite{DBLP:conf/kbse/LiL000J21,DBLP:conf/icsm/YuYCLZ22,DBLP:conf/nips/HashimotoGOL18,DBLP:conf/kbse/WeiLLXJ20,DBLP:conf/emnlp/ShiW0DZHZ022}, and AI-for-Science~\cite{wang2022retrieval}.

In the image domain, several studies~\cite{chen2022re,sheynin2022knn,blattmann2022retrieval,rombach2022text} employ cross-attention mechanisms to fuse retrieval results by integrating their latent representations. 
Conversely, Li et al.~\cite{li2022memory} implement a text-image Affine Combination Module (ACM) that directly concatenates hidden features.

Within the knowledge domain, several studies~\cite{DBLP:conf/naacl/OguzCKPOSGMY22,DBLP:conf/iclr/YuZNZL0HWWX23,DBLP:conf/cikm/DongLWZXX23,DBLP:journals/corr/abs-2308-13259,DBLP:conf/sigir/YuY23} have adopted FiD and its derivatives for downstream tasks. EaE~\cite{EaE} enhances the generator's understanding through entity-specific parameterization, while TOME~\cite{TOME} pivots to a nuanced encoding of mentions, prioritizing the granularity of mentions over entity representations alone.

In the field of 3D generation, ReMoDiffuse~\cite{DBLP:conf/iccv/ZhangGPCHLYL23} introduces a semantics-modulated attention mechanism which enhances the accuracy of generating corresponding 3D motions based on textual descriptions. AMD~\cite{jing2023amd} achieves efficient conversion from text to 3D motion by fusing the original diffusion process with the reference diffusion process.

In the audio domain, Koizumi et al.~\cite{DBLP:journals/corr/abs-2012-07331} utilized an LLM, incorporating encoded dense features in the attention module to guide the generation of audio captions. Re-AudioLDM~\cite{DBLP:journals/corr/abs-2309-08051} utilizes distinct encoders to extract deep features from text and audio, which are then integrated into the attention mechanism of its Latent Diffusion Model (LDM).

For video captioning, R-ConvED~\cite{DBLP:journals/tomccap/ChenPLYCM23} uses a convolutional encoder-decoder network to process retrieved video-sentence pairs with an attention mechanism, generating hidden states to produce captions. CARE~\cite{DBLP:journals/tip/YangCZ23} introduces a concept detector to produce concept probabilities, and incorporates concept representations into a hybrid attention mechanism. 

EgoInstructor~\cite{DBLP:journals/corr/abs-2401-00789} uses gated-cross attention to merge text and video features, improving the relevance and coherence of captions for egocentric videos.
Latent representation-based RAG, adaptable across modalities and tasks, blends retriever and generator hidden states but requires additional training for aligning latent spaces. It enables the development of sophisticated algorithms that seamlessly incorporate retrieved information.

\subsubsection{\textbf{Logit-based RAG}}

In logit-based RAG, generative models integrate retrieval information through logits during the decoding process. 
Typically, the logits are combined through simple summation or models to compute the probabilities for step-wise generation. 

In the text domain, kNN-LM~\cite{KNN-LM} and its variant~\cite{Efficient-KNNLM} blend language model probabilities with those from retrieval distances of similar prefixes at each decoding step.
TRIME~\cite{TRIME} and NPM~\cite{NPM} are radical evolutions of traditional kNN-LM approaches, using closely aligned tokens from a local database as output, particularly boosting performance in long-tail distribution scenarios.

Beyond text, other modalities, such as code and image, also leverage logit-based RAG. 

In the domain of code, several studies~\cite{DBLP:conf/icse/ZhangW00020,DBLP:conf/emnlp/Zhang0YC23} have also adopted the concept kNN to enhance final output control, thereby achieving superior performance. 
Furthermore, EDITSUM~\cite{DBLP:conf/kbse/LiL000J21} improves the quality of code summarization by integrating prototype summaries at the logit level.
For image captioning, MA~\cite{fei2021memory} directly applies the kNN-LM framework to address the image caption problem, achieving favorable results.
In summary, logit-based RAG utilizes historical data to deduce current states and merges information at the logit level, ideal for sequence generation. It focuses on generator training and allows for novel methods that capitalize on probability distributions for future tasks.

\subsubsection{\textbf{Speculative RAG}}
Speculative RAG seeks opportunities to use retrieval instead of pure generation, aiming to save resources and accelerate response speed.
REST~\cite{REST} replaces the small models in speculative decoding~\cite{Speculative_Decoding} with retrieval, enabling the generation of drafts.
GPTCache~\cite{GPTCache} addresses the issue of high latency when using the LLM APIs by building a semantic cache for storing LLM responses.
COG~\cite{COG} decomposes the text generation process into a series of copy-and-paste operations, retrieving words or phrases from the documents instead of generation.
Cao et al.~\cite{RetrievalisAccurateGeneration} proposed a new paradigm to eliminate the dependence of the final result on the quality of the first-stage retrieved content, replacing generation with directly retrieved phrase level content. 

In conclusion, speculative RAG is currently primarily applicable to sequential data. 
It decouples the generator and the retriever, enabling the direct use of pre-trained models as components. 
Within this paradigm, we can explore a wider range of strategies to effectively utilize the retrieved content.

\subsection{\textbf{RAG Enhancements}}
\label{sec:enhancements}

\begin{figure*}
\centering
\includegraphics[width=0.9\textwidth]{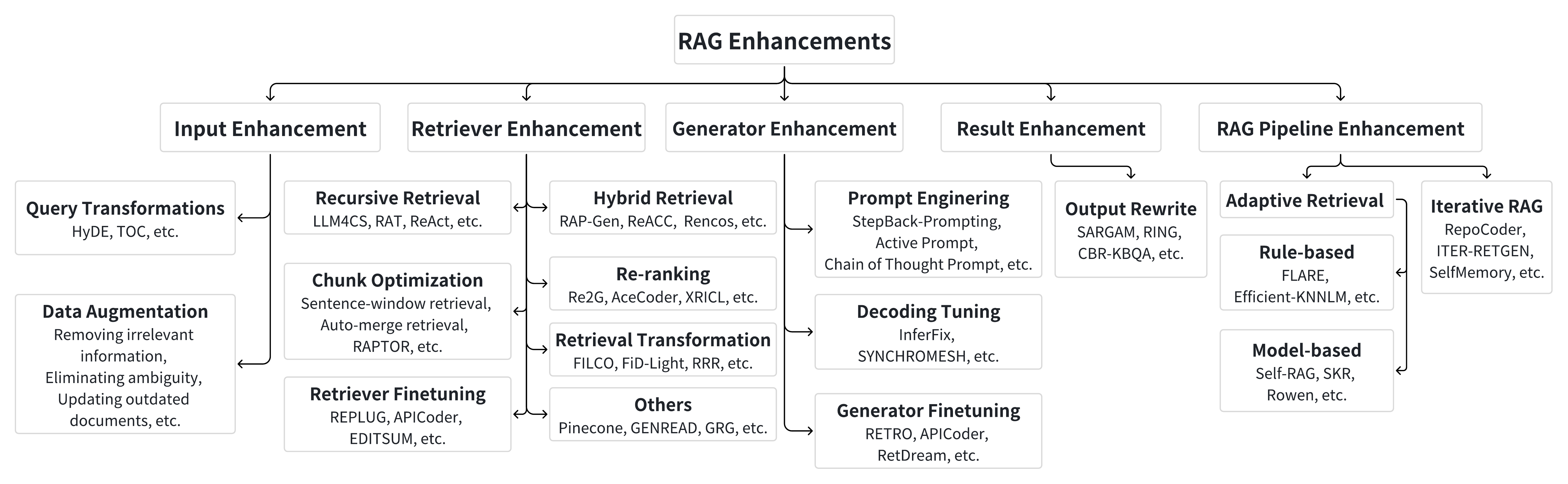}
\caption{Taxonomy of RAG Enhancements.}
\label{fig:Improvement}

\end{figure*}

In this section, we introduce methods which enhance the performance of a constructed RAG system.
We categorize existing methods into 5 groups based on their enhancement targets: input, retriever, generator, result, and the entire pipeline.

\subsubsection{\textbf{Input Enhancement}}

The input, initially fed into the retriever, significantly impacts the final outcome of the retrieval stage. In this section, we introduce two methods for input enhancement: query transformation and data augmentation.

\noindent\textbf{Query Transformation:}
Query transformation can enhance the result of retrieval by modifying the input query.

Query2doc~\cite{Query2doc} and HyDE~\cite{HyDE} use the original query to generate a pseudo document, which is later used as the query for retrieval. The pseudo document contains richer relevant information, which helps to retrieve more accurate results.
TOC~\cite{TOC} leverages retrieved contents to decompose the ambiguous query into multiple clear sub-queries, which are sent to the generator and aggregated to produce the final result.
For complex or ambiguous queries, RQ-RAG~\cite{RQ-RAG} breaks them down into clear subqueries for fine-grained retrieval and synthesizes the responses to deliver a cohesive answer to the original query.
Tayal et al.~\cite{tayal2024dynamic} refined the initial query using dynamic few-shot examples and context retrieval, enhancing the generator's grasp of user intent.

\noindent\textbf{Data Augmentation:}
Data augmentation improves data before retrieval, including techniques such as removing irrelevant information, eliminating ambiguity, updating outdated documents, synthesize new data, etc.

Make-An-Audio~\cite{DBLP:conf/icml/HuangHY0LLYLYZ23} uses captioning and audio-text retrieval to generate captions for language-free audio to mitigate data sparsity, and adds random concept audio to improve the original audio.
LESS~\cite{LESS} optimizes dataset selection for downstream tasks by analyzing gradient information, aiming to enhance model performance in response to instructional prompts.
ReACC~\cite{DBLP:conf/acl/LuDHGHS22} employs data augmentation (including renaming and dead code insertion) to pre-train the code retrieval model.
Telco-RAG~\cite{Telco-RAG} enhances the retrieve accuracy by appling a ``Vocabulary for 3GPP Specifications'', and match them to user queries with a router module.
\subsubsection{\textbf{Retriever Enhancement}}

In RAG systems, the quality of retrieved content determines the information fed into the generators. Lower content quality increases the risk of model hallucinations or other degradation. In this section, we introduce efficient ways to enhance retrieval effectiveness.

\noindent\textbf{Recursive Retrieval:}
Recursive retrieval is to perform multiple searches to retrieve richer and higher-quality contents.

ReACT~\cite{ReAct} uses Chain-of-Thought (CoT)~\cite{COT} to break queries down for recursive retrieval and provide richer information.
RATP~\cite{RATP} uses the Monte-Carlo Tree Search for simulations to select optimal retrieval content, which is then templated and forwarded to the generator for output.

\noindent\textbf{Chunk Optimization:}
Chunk optimization refers to adjusting chunk size for improved retrieval results.

LlamaIndex~\cite{LlamaIndex} incorporates a series of chunk optimization methods, one of which operates on a `small to big' principle. The core concept here is to pinpoint finer-grained content but return richer information. 
For instance, Sentence-window retrieval fetches small text chunks and returns a window of relevant sentences surrounding the retrieved segment.
In auto-merge retrieval, documents are arranged in a tree structure. The process retrieves the parent node, which encapsulates the content of its child nodes, by fetching the child node first.
To address the lack of contextual information, RAPTOR~\cite{RAPTOR} employs recursive embedding, clustering, and summarization of text chunks until further clustering becomes infeasible, thereby constructing a multi-level tree structure.
Prompt-RAG~\cite{Prompt-RAG} enhances retrieval accuracy by pre-generating a table of contents, enabling the model to autonomously select relevant chapters based on the query.
Raina et al.~\cite{raina2024question} break text chunks into finer atomic statements to achieve higher recall and improved results.

\noindent\textbf{Retriever Finetuning:}
The retriever, central to the RAG system, relies on a proficient embedding model~\cite{bge_embedding,bge_m3,cocktail,llm_embedder} to represent related content and feed the generator, enhancing system performance.

Additionally, embedding models with strong expressive power can be fine-tuned with domain-specific or task-related data to boost performance in targeted areas.
REPLUG~\cite{REPLUG} treats LM as a black box and update the retriever model based on the final results. APICoder~\cite{DBLP:conf/emnlp/ZanCLGWL22} finetunes the retriever with python files and api names, signature, description. 
EDITSUM~\cite{DBLP:conf/kbse/LiL000J21} finetunes the retriever to decrease the jaccard distance between summaries after retrieval.
SYNCHROMESH~\cite{DBLP:conf/iclr/PoesiaP00SMG22} adds tree distance os ASTs in the loss and uses Target Similarity Tuning (TST) to finetune the retriever.
R-ConvED~\cite{DBLP:journals/tomccap/ChenPLYCM23} finetunes the retriever with the same data as generator.
Kulkarni et al.~\cite{RL4RAG} applied infoNCE loss to finetune the retriever.

\noindent\textbf{Hybrid Retrieval:}
Hybrid retrieve denotes the concurrent employment of a diverse array of retrieval methodologies or the extraction of information from multiple distinct sources.

RAP-Gen~\cite{DBLP:conf/sigsoft/Wang0JH23}, BlendedRAG~\cite{Blended-RAG}and ReACC~\cite{DBLP:conf/acl/LuDHGHS22} use both dense retriever and sparse retriever to improve the quality of retrieval. Rencos~\cite{DBLP:conf/icse/ZhangW00020}
uses sparse retriever to retrieve similar code snippets on syntactic-level and uses dense retriever to retrieve similar code snippets on semantic-level. 
BASHEXPLAINER~\cite{DBLP:conf/icsm/YuYCLZ22} first uses dense retriever to capture semantic information and then uses sparse retriever to acquire lexical information.
RetDream~\cite{DBLP:journals/corr/abs-2402-02972} first retrieves with text and then retrieves with the image embedding.
CRAG~\cite{CRAG} features a retrieval evaluator that gauges document relevance to queries, prompting three retrieval responses based on confidence: direct use of results for Knowledge Refinement if accurate, Web Search if incorrect, and a hybrid approach for ambiguous cases.
Huang et al.~\cite{RAGAE} improved question-answering by introducing DKS (Dense Knowledge Similarity) and RAC (Retriever as Answer Classifier) in the retrieval phase, evaluating answer relevance and knowledge applicability.
UniMS-RAG~\cite{UniMS-RAG} introduces a novel kind of token, termed as the ``acting token'', which determines the source from which to retrieve information.
Koley et al.~\cite{koley2024you} enhance image retrieval by integrating sketch and text for fine-grained retrieval, yielding improved results.

\noindent\textbf{Re-ranking:}
The Rerank technique refers to reordering the retrieved content in order to achieve greater diversity and better results.

Re2G~\cite{Re2G} applies a re-ranker~\cite{ReRanker} model after the traditional retriever to reduce the impact of information loss caused by compressing text into vectors.
AceCoder~\cite{li2023acecoder} reranks the retrieved programs with a selector to reduce redundant programs and obtain diverse retrieved programs. 
XRICL~\cite{DBLP:conf/emnlp/0010Z0L22} uses a distillation-based exemplar reranker after retrieval.
Rangan~\cite{rangan2024fine} employs the Quantized Influence Measure, assessing statistical biases between a query and a reference to evaluate the similarity of data subsets and rerank retrieval results.
UDAPDR~\cite{UDAPDR} uses LLMs to cost-effectively generate synthetic queries that train domain-specific rerankers, which then apply multi-teacher knowledge distillation to develop a cohesive retriever.
LLM-R~\cite{LLM-R} refines its retriever iteratively by employing a static LLM for document ranking and reward model training, complemented by knowledge distillation. Each training cycle incrementally improves the retriever, enabling progressive optimization.
Finardi et al.~\cite{finardi2024chronicles} integrated reciprocal rank into the retrieval process for enhanced text chunk relevance, and utilized monoT5 as a reranker to optimize the result quality.
Li et al.~\cite{li2024enhancing} integrate a reranking module into their end-to-end RAG system, enhancing the retrieval quality and factual accuracy of LLMs.

\noindent\textbf{Retrieval Transformation:}
Retrieval Transformation involves rephrasing retrieved content to better activate the generator's potential, resulting in improved output.

FILCO~\cite{FILCO} efficiently purges extraneous material from retrieved text, isolating only the pertinent supporting content to streamline the generator's task and facilitate accurate answer prediction.
FiD-Light~\cite{FiD-Light} initially employs an encoder to convert the retrieved content into a vector, which it then compresses, resulting in a substantial reduction of latency time.
RRR~\cite{RRR} integrates the current query with the top-k document in each round through a template, and subsequently restructures it via a pre-trained LLMs (GPT-3.5-Turbo etc.).

\noindent\textbf{Others:}
In addition to the above optimization methods, there are also some other optimization methods for the retrieve process. 

For example, meta-data filtering ~\cite{Pinecone} is a method to help processing retrieved documents which uses metadata (such as time, purpose, etc.) to filter the retrieved documents for better results. GENREAD~\cite{GENREAD} and GRG~\cite{GRG} introduce a novel approach where the retrieval process is supplanted or improved by prompting a LLM to generate documents in response to a given question. Multi-Head-RAG~\cite{Multi-Head-RAG} employs multiple embedding models to project the same text chunk into various vector spaces and utilizes a multi-head attention layer to capture different informational aspects, thereby increasing the accuracy of the retrieval process.

\subsubsection{\textbf{Generator Enhancement}}
In RAG systems, the quality of the generator often determines the quality of the final output results. Therefore, the ability of the generator determines the upper limit of the entire RAG system's effectiveness. 

\noindent\textbf{Prompt Engineering:}
Technologies in prompt engineering~\cite{Prompt_Engineering_Guide} that focus on improving the quality of LLMs' output, such as prompt compression, Stepback Prompt~\cite{StepBack-Prompting}, Active Prompt~\cite{active-prompt}, Chain of Thought Prompt~\cite{COT}, etc., are all applicable to LLM generators in RAG systems.

LLMLingua~\cite{LLMLingua} applies a small model to compresses the overall length of the query to accelerate model inference, relieving the negative impact of irrelevant information on the model and alleviating the phenomenon of ``Lost in the Middle''\cite{Lost_in_the_middle}.
ReMoDiffuse~\cite{DBLP:conf/iccv/ZhangGPCHLYL23} decomposes  complex descriptions into anatomical text scripts by using ChatGPT.
ASAP~\cite{ahmed2024automatic} incorporates exemplar tuples, consisting of input code, function definitions, analysis results, and corresponding comments, into prompts to yield better results.
CEDAR~\cite{DBLP:conf/icse/NashidSM23} uses a designed prompt template to organize code demonstration, query, and natural language instructions into a prompt.
XRICL~\cite{DBLP:conf/emnlp/0010Z0L22} utilizes COT technology to add translation pairs as an intermediate step in cross linguistic semantic parsing and inference.
ACTIVERAG~\cite{ACTIVERAG} employs the Cognition Nexus mechanism to  calibrate the intrinsic cognition of LLMs and applies COT prompt in answer generation.
Make-An-Audio~\cite{DBLP:conf/icml/HuangHY0LLYLYZ23} is able to use other modalities as input which can provide much richer information for the following process.

\noindent\textbf{Decoding Tuning:}
Decoding tuning involves enhancing generator control by fine-tuning hyperparameters for increased diversity and constraining the output vocabulary, among other adjustments.

InferFix~\cite{DBLP:conf/sigsoft/JinSTSLSS23} balances the diversity and quality of results by adjusting the temperature in decoder.
SYNCHROMESH~\cite{DBLP:conf/iclr/PoesiaP00SMG22} limits the output vocabulary of the decoder by implementing a completion engine to eliminate implementation errors.

\noindent\textbf{Generator Finetuning:}
The finetuning of the generator can enhance the model's ability to have more precise domain knowledge or better fit with the retriever. 

RETRO~\cite{RETRO} fixes the parameters of the retriever and uses the chunked cross attention mechanism in the generator to combine the content of the query and retriever. 
APICoder~\cite{DBLP:conf/emnlp/ZanCLGWL22} finetunes the generator CODEGEN-MONO 350M~\cite{CODEGEN-MONO} with a shuffled new file combined with API information and code blocks. 
CARE~\cite{DBLP:journals/tip/YangCZ23} trains encoders with image, audio, and video-text pairs, then fine-tunes the decoder (generator) to simultaneously reduce caption and concept detection loss, while keeping the encoders and retriever fixed.
Animate-A-Story~\cite{DBLP:journals/corr/abs-2307-06940} optimizes the video generator with image data, and then finetunes a LoRA~\cite{LoRA} adapter to capture the appearance details of the given character.
RetDream~\cite{DBLP:journals/corr/abs-2402-02972} finetunes a LoRA adapter~\cite{LoRA} with the rendered images.

\subsubsection{\textbf{Result Enhancement}}
In many scenarios, the result of RAG may not achieve the expected effect, and some techniques of Result Enhancement can help alleviate this problem.

\noindent\textbf{Output Rewrite:}
Output Rewrite refers to rewriting the content generated by the generator in certain scenarios to meet the needs of downstream tasks.
SARGAM~\cite{DBLP:journals/corr/abs-2306-06490} refines outputs in code-related tasks by employing a special Transformer alongside Deletion, Placeholder, and Insertion Classifiers to better align with the real-world code context. 
Ring~\cite{DBLP:conf/aaai/JoshiSG0VR23} obtains diversity results by reranking candidates based on the average of per token log probabilities produced by the generator. 
CBR-KBQA~\cite{DBLP:conf/emnlp/DasZTGPLTPM21} revises the result by aligning generated relations with those presented in the local neighborhood of the query entity in knowledge graph.
\subsubsection{\textbf{RAG Pipeline Enhancement}}
RAG pipeline enhancement refers to optimizing the overall process of RAG in order to achieve better performance results.

\noindent\textbf{Adaptive Retrieval:}
Some studies on RAG suggest that retrieval doesn't always enhance the final results. Over-retrieval can lead to resource wastage and potential confusion when the model's inherent parameterized knowledge suffices for answering relevant questions. Consequently, this chapter will delve into two methods for determining retrieval necessity: rule-based and model-based approaches.

\textit{Rule-based:} FLARE~\cite{FLARE} actively decides whether and when to search through the probability in the generation process. Efficient-KNNLM~\cite{Efficient-KNNLM} combines the generation probability of KNN-LM~\cite {KNN-LM} and NPM~\cite {NPM} with a hyperparameter $\lambda$ to determine the proportion of generation and retrieval. Mallen et al.~\cite{Adaptive-Retrieval-whennottrust} used statistical analysis on questions to enable direct answers for high-frequency ones and applied RAG for low-frequency ones.
Jiang et al.~\cite{lm-calibration} evaluated model confidence based on Model Uncertainty, Input Uncertainty, and Input Statistics to guide retrieval decisions.
Kandpal et al.~\cite{LLM_Struggle_to_Learn_Long-Tail_Knowledge} studied the correlation between the number of relevant documents and the model's knowledge mastery to assess the need for retrieval.

\textit{Model-based:} 
Self-RAG~\cite{Self-RAG} uses a trained generator to determine whether to perform a retrieval based on the retrieve token under different user querys.
Ren et al.~\cite{LLM-Knowledge-Boundary} used ``Judgment Prompting'' to determine whether LLMs can answer relevant questions and whether their answers are correct or not, thereby assisting in determining the necessity of a retrieval. 
SKR~\cite{SKR} uses the ability of LLMs themselves to judge in advance whether they can answer the question, and if they can answer, no retrieval is performed. 
Rowen~\cite{Rowen} translates a question into multiple languages and checks for answer consistency across these languages, using the results to determine the need for information retrieval.
AdaptiveRAG~\cite{AdaptiveRAG} dynamically decides whether to retrieve based on the query complexity by a classifier, which is a smaller LM.

\noindent\textbf{Iterative RAG:}
Iterative RAG progressively refines results by repeatedly cycling through retrieval and generation phases, rather than a single round.

RepoCoder~\cite{DBLP:conf/emnlp/ZhangCZKLZMLC23} uses an iterative retrieval-generation approach for code completion, refining queries with previously generated code to better utilize dispersed information and improve outcomes.
ITER-RETGEN~\cite{ITER-RETGEN} iteratively enhances content quality by using the generator's output to pinpoint knowledge gaps, retrieving necessary information, and informing future generation cycles.
SelfMemory~\cite{SelfMemory} utilizes a retrieval-augmented generator iteratively to form an expansive memory pool, from which a memory selector picks an output to inform the next generation cycle.
RAT~\cite{RAT} initially generates content by an LLM with a zero-shot CoT prompt, then revises each thought step by retrieving knowledge from external knowledge base.

\section{\textbf{Applications}}
\label{sec:applications}

\begin{table*}[!t] 
  \centering
  \caption{Taxonomy of RAG applications across various modalities.}
  \label{tb:applications}
\small

\begin{adjustbox}{width=1\textwidth,center}
\small
\begin{tabular}{|c|c|c|c|c|}
\hline
\multicolumn{5}{|c|}{\textbf{RAG for Text}}\\ \hline
\multicolumn{1}{|c|}{\textbf{Question Answering}} &
  \multicolumn{1}{c|}{\textbf{Human-Machine Conversation}} &
  \multicolumn{1}{c|}{\textbf{Neural Machine Translation}} &
  \multicolumn{1}{c|}{\textbf{Summarization}} &
   \multicolumn{1}{c|}{\textbf{Others}} \\ \hline
  \begin{tabular}[c]{@{}l@{}} 
  \small
  \colorbox{c1}{ REALM\two \three}  \colorbox{c1}{ TKEGEN\three}    \colorbox{c4}{  RIAG\two}\\ \colorbox{c2}{ Fid\two \three} \colorbox{c2}{ RETRO\three} \colorbox{c3}{ NPM\two \three} \\ \colorbox{c1}{ SKR\three \five} \colorbox{c1}{ Self-RAG\three \five} \colorbox{c1}{ TOG\two}\end{tabular} &
  \begin{tabular}[c]{@{}l@{}}\colorbox{c2}{ ConceptFlow\two \three} \colorbox{c2}{ Skeleton-to-Response\two \three} \\  \colorbox{c1}{ CREA-ICL\one \two}\colorbox{c2}{ Internet-Augmented-DG\two \three} \\\colorbox{c1}{ BlenderBot3\two \three}  \colorbox{c1}{ CEG\two \four}\end{tabular} &
  \begin{tabular}[c]{@{}l@{}} \colorbox{c2}{ NMT-with-Monolingual-TM\one \two \three} \\ \colorbox{c3}{ KNN-MT\two \three} \colorbox{c4}{  COG\two} \colorbox{c3}{ TRIME\two \three} \end{tabular} &
  \begin{tabular}[c]{@{}l@{}}\colorbox{c1}{ RAMKG\two \three}  \colorbox{c2}{ Unlimiformer\three} \\ \colorbox{c1}{ RPRR\two} \colorbox{c1}{ RIGHT\two \three}\end{tabular} &
  \begin{tabular}[c]{@{}l@{}}\colorbox{c1}{ CONCRETE\two \three}  \colorbox{c1}{ Atlas\two \three}\\  \colorbox{c2}{ KG-BART\two \three}  \colorbox{c1}{ R-GQA\two \three}\end{tabular} \\ \hline
\end{tabular}
 \end{adjustbox}

\begin{adjustbox}{width=1\textwidth,center}
\large
\begin{tabular}{|c|c|c|c|c|c|}
\hline
\multicolumn{6}{|c|}{\textbf{RAG for Code}}\\ \hline
\textbf{Code Generation} &
\textbf{Code Summary} & 
\textbf{ Code Completion } &
\textbf{ Automatic Program Repair} &
\makecell{\textbf{Text-to-SQL and}\\ \textbf{ Code-based Semantic Parsing}} & 
\textbf{ Others} \\
\hline

\begin{tabular}[c]{@{}l@{}}
\colorbox{c1}{ SKCODER\three}  \colorbox{c1}{  RRGCode\two}\\ \colorbox{c1}{  ARKS\one \five} \colorbox{c3}{  KNN-TRANX\four}\\  \colorbox{c3}{  RECODE} \colorbox{c4}{   Toolcoder\three \four}\end{tabular} &
  \begin{tabular}[c]{@{}l@{}}\colorbox{c2}{ RACE\one}   \colorbox{c2}{  BASHEXPLAINER\two}\\ \colorbox{c2}{ READSUM\four}  \colorbox{c3}{  Rencos\two}\\ \colorbox{c3}{  CoRec\two} \colorbox{c3}{Tram\three}   \GradientTwo{EDITSUM\two} \end{tabular} &
  \begin{tabular}[c]{@{}l@{}}\colorbox{c1}{  ReACC\one \two} \colorbox{c1}{  RepoCoder\one \three \five}\\ \colorbox{c1}{  De-Hallucinator\five} \colorbox{c1}{REPOFUSE\three}\\ \colorbox{c2}{RepoFusion\three} \colorbox{c2}{EDITAS\three}\end{tabular} &
  \begin{tabular}[c]{@{}l@{}}\colorbox{c1}{ RING\four} \colorbox{c1}{ CEDAR\three}\\ \colorbox{c1}{ RAP-Gen\two \three} \colorbox{c1}{ InferFix\three}\\ \colorbox{c1}{ SARGAM\three} \colorbox{c1}{ RTLFixer\two \three}\end{tabular} &
  \begin{tabular}[c]{@{}l@{}}\colorbox{c1}{ XRICL\two \three} \colorbox{c1}{ SYNCHROMESH\two \three}\\ \colorbox{c1}{ RESDSQL\three} \colorbox{c1}{ REFSQL\two \three}\\ \colorbox{c1}{ CodeICL\three} \colorbox{c1}{ MURRE\four \five}\end{tabular} &
  \begin{tabular}[c]{@{}l@{}}\colorbox{c1}{ StackSpotAI\two \three} \colorbox{c1}{ E\&V}\\ \colorbox{c1}{ Code4UIE\three} \colorbox{c1}{ De-fine\two \four}\\ \colorbox{c1}{ ImputBlaster\five}\end{tabular} \\ \hline
\end{tabular}
 \end{adjustbox} 

{
\renewcommand{\arraystretch}{1.2}
\begin{adjustbox}{width=1.0\textwidth,center}
\tiny
    \centering

\begin{tabular}{|c|c|c|c|}

\hline
\multicolumn{4}{|c|}{\textbf{RAG for Knowledge}}\\ \hline
\textbf{\makecell{Knowledge Base QA}} &
\textbf{\makecell{Knowledge-augmented Open-domain QA}} &
\textbf{\makecell{Table for QA}} &
\textbf{Others} \\ \hline
\begin{tabular}[c]{@{}l@{}}\colorbox{c1}{ CBR-KBQA\two \three \four}  \colorbox{c1}{ TIARA\one \two \three} \colorbox{c1}{ Keqing\one \two \three}\\ \colorbox{c1}{ RNG-KBQA\two \four} \colorbox{c2}{ ReTraCk\three} \colorbox{c2}{ SKP\one \two \three}\end{tabular} &
  \begin{tabular}[c]{@{}l@{}}\colorbox{c2}{ UniK-QA\one \two} \colorbox{c2}{ KG-FiD\two} \colorbox{c2}{ GRAPE\two}  \\ \colorbox{c2}{ SKURG\one \two} \colorbox{c1}{ KnowledGPT\two} \colorbox{c1}{ EFSUM\three}\end{tabular} &
  \begin{tabular}[c]{@{}l@{}}\colorbox{c2}{ EfficientQA\two}  \colorbox{c2}{ CORE\three}  \colorbox{c2}{ Convinse\one \two} \\ \colorbox{c2}{ RINK\two \three} \colorbox{c1}{ T-RAG\two \three}   \colorbox{c1}{ StructGPT\two}\end{tabular} &
  \begin{tabular}[c]{@{}l@{}}\colorbox{c1}{ GRetriever\three} \colorbox{c1}{ SURGE\three} \\\colorbox{c1}{ K-LaMP} \colorbox{c2}{ RHO\four}\end{tabular} \\ \hline
 \end{tabular}

 \begin{tabular}{|c|}
      \hline
      \multicolumn{1}{|c|}{\textbf{RAG for 3D}}\\ \hline
      \textbf{\makecell{Text-to-3D }} \\ \hline
  \begin{tabular}[c]{@{}l@{}}\colorbox{c2}{ ReMoDiffuse\one \two} \\
  \colorbox{c1}{ AMD\one}  \end{tabular} \\ \hline
      \end{tabular}
\end{adjustbox}
}

\begin{adjustbox}{width=1.0\textwidth,center}
\footnotesize
\begin{tabular}{|c|c|c|}
\hline
\multicolumn{3}{|c|}{\textbf{RAG for Image}}\\ \hline
\textbf{Image Generation} &
  \textbf{Image Captioning} &
  \textbf{Others} \\ \hline
\begin{tabular}[c]{@{}l@{}}\colorbox{c1}{ RetrieveGAN\two} \colorbox{c1}{ IC-GAN\three}   \colorbox{c2}{ Re-imagen\three}  \\ \colorbox{c2}{ RDM} \colorbox{c2}{ Retrieve\&Fuse\three} \colorbox{c2}{ KNN-Diffusion}\end{tabular} &
  \begin{tabular}[c]{@{}l@{}}\colorbox{c3}{ MA\four}  \colorbox{c1}{ REVEAL\two}  \colorbox{c1}{ SMALLCAP\one} \\ \colorbox{c1}{ CRSR\one}  \colorbox{c2}{ RA-Transformer}\end{tabular} &
  \begin{tabular}[c]{@{}l@{}}\colorbox{c1}{ PICa\four} \colorbox{c1}{ Maira\two}\\ \colorbox{c1}{ KIF\two} \colorbox{c1}{ RA-VQA\two}\end{tabular} \\ \hline
\end{tabular}

\begin{tabular}{|c|c|c|}
\hline
\multicolumn{3}{|c|}{\textbf{RAG for Video}}\\ \hline
\textbf{Video Captioning} &
  \textbf{Video QA\&Dialogue} &
  \textbf{Others} \\ \hline
\begin{tabular}[c]{@{}c@{}}\colorbox{c3}{ KaVD\two \three} \colorbox{c2}{ R-ConvED\two \three}  \\ 
\colorbox{c2}{\strut CARE\three}\GradientText{{EgoInstructor\one \two \three}}\end{tabular} &
  \begin{tabular}[c]{@{}c@{}}\colorbox{c2}{ MA-DRNN\one \two}  \colorbox{c1}{ R2A\two}  \\ \colorbox{c2}{ Tvqa+\three} \colorbox{c1}{ VGNMN\two}\end{tabular} &
  \begin{tabular}[c]{@{}c@{}}\colorbox{c1}{ VidIL\one \two} \colorbox{c1}{ RAG-Driver\two} \\ \colorbox{c2}{ Animate-A-Story\one \three}\end{tabular} \\ \hline
\end{tabular}
\end{adjustbox}

\begin{adjustbox}{width=1.0\textwidth,center}
\footnotesize

\begin{tabular}{|c|c|c|}
\hline
\multicolumn{3}{|c|}{\textbf{RAG for Science}}\\ \hline
\textbf{Drug Discovery} & \textbf{Biomedical Informatics Enhancement}      & \textbf{Math Applications} \\ \hline
\colorbox{c1}{ RetMol\one \three} \colorbox{c2}{ PromptDiff\one \two}        & \colorbox{c4}{  PoET\two}  \colorbox{c1}{ Chat-Orthopedist\one \three} \colorbox{c1}{ BIOREADER\one} \colorbox{c1}{ MedWriter\two} \colorbox{c1}{ QARAG\one \two} & \colorbox{c4}{  LeanDojo\two}  \colorbox{c1}{ RAG-for-math-QA\one \two}   \\ \hline
\end{tabular}

 \begin{tabular}{|c|c|}
\hline
\multicolumn{2}{|c|}{\textbf{RAG for Audio}}\\ \hline
\textbf{\makecell{Audio Generation}}  & \textbf{\makecell{Audio Captioning}} \\ \hline
\colorbox{c2}{Re-AudioLDM\three}   \colorbox{c1}{ Make-An-Audio\one \three} & \colorbox{c1}{ RECAP\two\three}                     \\ \hline
\end{tabular}

\end{adjustbox}

    \begin{minipage}[t]{0.4\textwidth}
    \scriptsize
    \centering
    \colorbox{c1}{Query-based}\quad 
    \colorbox{c2}{Latent-based}\quad 
    \colorbox{c3}{Logit-based}\quad 
    \\
    \colorbox{c4}{Speculative} 
      \GradientText{Query+Latent}
      \GradientTwo{{Latent+Logit}}
  \end{minipage}%
  \begin{minipage}[t]{0.4\textwidth}
  \scriptsize
    \centering
    \colorbox{white}{$\dagger$  Input}\quad 
    \colorbox{white}{$\ddagger$  Retriever}\quad 
    \colorbox{white}{ $\mathsection$  Generator}\quad 
    \\
    \colorbox{white}{$\|$  Output}\quad 
    \colorbox{white}{$\mathparagraph$  Pipeline}\quad 
  \end{minipage}
\end{table*}

In this section, we focus on RAG applications spanning various modalities. To echo with the taxonomy of RAG foundations and enhancements, we also demonstrate their utilization across different tasks in Table~\ref{tb:applications}. 
\subsection{\textbf{RAG for Text}}

To begin with, text generation is among the most important and widely deployed applications for RAG. Here we introduce popular works for seven tasks, respectively.
\subsubsection{\textbf{Question Answering}}

\begin{figure}[h]
  \centering
  \includegraphics[width=\linewidth]{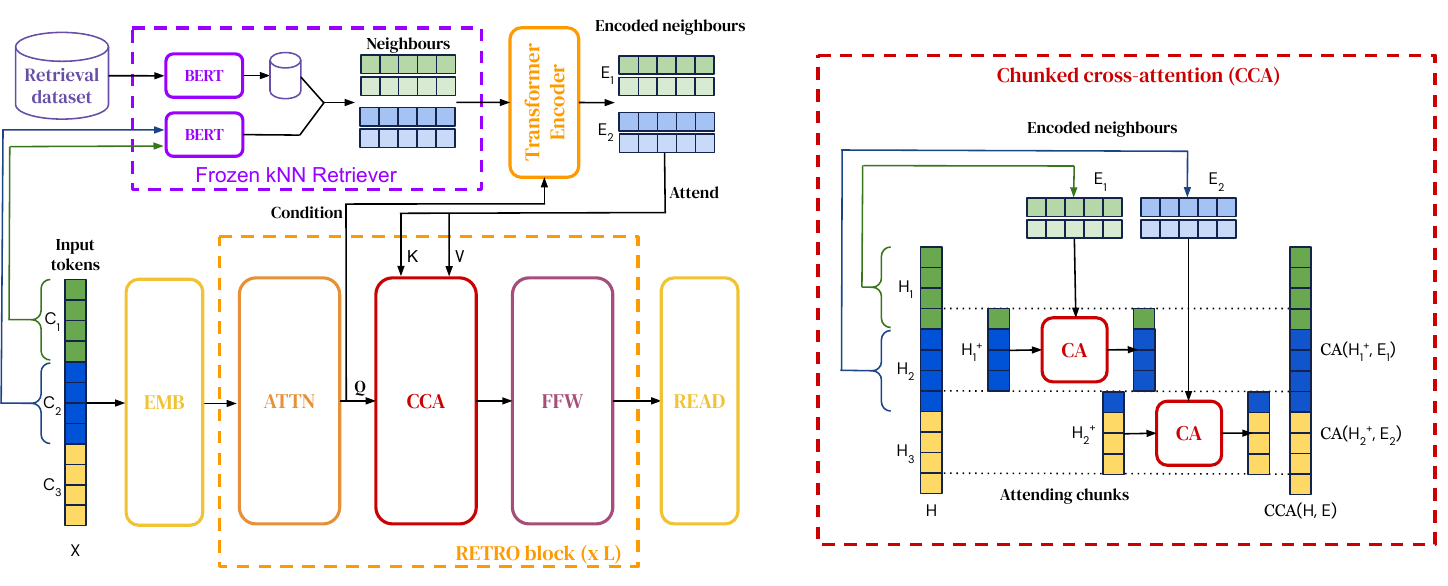}
  \caption{Architecture of RETRO~\cite{RETRO} model.}
  \label{fig:retro}
\end{figure}

Question answering involves the process of providing responses to posed questions by drawing from a vast and comprehensive collection of textual sources.
FiD~\cite{FId} and REALM~\cite{REALM} identify the top-k most pertinent article snippets based on the query and forward each snippet along with the question to LLMs to generate k responses. These responses are then synthesized into a final answer. 
Toutanova et al.~\cite{TKEGEN} substituted the text corpus in REALM with subgraphs from a knowledge graph, yielding impressive results. 
As shown in Fig.~\ref{fig:retro}, RETRO~\cite{RETRO} employs attention mechanisms to integrate the question with relevant retrieved documents within the model to produce the final answer. 
SKR~\cite{SKR} observes that using RAG does not invariably benefit question answering and thus explored guiding the model to evaluate its grasp of pertinent knowledge, subsequently adapting its use of external resources for retrieval enhancement.
TOG~\cite{TOG} introduces an innovative knowledge graph-augmented LLM framework, which excels by fostering interactions between LLMs and the knowledge graph and by expanding the inference path space with beam search.
NPM~\cite{NPM} pioneers the use of nonparametric data distributions in lieu of the softmax layer, enabling models with fewer parameters to perform effectively. 
CL-ReLKT~\cite{CL-ReLKT} employs a language-generalized encoder to bridge the gap between question-document pairs across languages, thus better leveraging multilingual data. CORE~\cite{CORE} mitigates language resource disparities by introducing a novel dense passage retrieval algorithm and a multilingual autoregressive generation model. Lastly, EAE~\cite{EaE} enhances answer quality by retrieving entity embeddings for query entities and integrating these with hidden states for further processing. 
UR-QA~\cite{UR-QA} proposes to simultaneously retrieve QA pairs and text chunks, selecting the final answer by comparing their calibrated confidences.
DISC-LawLLM~\cite{DISC-LawLLM} constructs a supervised fine-tuning dataset through a legal syllogism prompting strategy, enabling the model to receive support from the latest legal information.
RAG-end2end~\cite{RAG-end2end} conducts simultaneous training of the retriever (DPR) and the generator (BART) to optimize performance for the end-to-end question-answering task and to facilitate domain adaptation.
MultiHop-RAG~\cite{MultiHop-RAG} extracts and aggregates information from distinct documents, providing the generator with the necessary context for definitive query answers.

\subsubsection{\textbf{Fact Verification}}
Fact verification typically refers to determining whether a given natural language text and a related claim or assertion match the facts in the text.

CONCRETE~\cite{CONCRETE} leverages cross-lingual retrieval mechanisms to tap into a wealth of multilingual evidence, effectively bridging the gap in resources for languages that are underrepresented in fact-checking datasets.
Atlas~\cite{Atlas} shows that using RAG to support LLMs in knowledge-intensive tasks markedly improves their few-shot learning performance.
Hagström et al.~\cite{hagstrom2023effect} proved on LLaMA~\cite{LLaMA} and Atlas~\cite{Atlas} that search augmentation is more beneficial for solving inconsistency problems than increasing model size.
Stochastic RAG~\cite{Stochastic_RAG} employs stochastic sampling without replacement to address the non-differentiable topk selection process in RAG retrieval, enabling end-to-end optimization and achieving excellent results in fact verification scenarios.
\subsubsection{\textbf{Commonsense Reasoning}}
Commonsense reasoning entails the capability of machines to infer or make decisions on problems or tasks in a human-like manner, drawing upon their acquired external knowledge and its application.

KG-BART~\cite{KG-BART} expands the conceptual landscape by incorporating intricate interrelations among diverse concepts within a knowledge graph. It employs graph attention mechanisms to aid LLMs in crafting more nuanced and logically coherent sentences. 
Wan et al.~\cite{CONFLICTINGQA} constructed the CONFLICTINGQA dataset with contentious questions and conflicting answers to study how textual features affect LMs' handling of controversial issues.

\subsubsection{\textbf{Human-Machine Conversation}}

Human-machine conversation encompasses the ability of machines to comprehend natural language and adeptly employ this skill to engage with humans seamlessly.

ConceptFlow~\cite{KG_enhanced_conversation} leverages a commonsense knowledge graph to structure conversations, directing the flow of dialogue based on attention scores, and propelling the conversation forward.
Cai et al.~\cite{Skeleton-to-Response} reimagined the text generation task as a cloze test by retrieving and distilling the essence of past conversational history, leading to notable outcomes. Komeili et al.~\cite{Internet-Augmented-DG} augmented dialogue generation quality by harnessing advanced search engine technologies to source pertinent content from the internet. BlenderBot3~\cite{BlenderBot3} broadens its search horizon, not only mining relevant internet content but also local dialogue history, and employs entity extraction among other techniques to refine the quality of the resulting dialogue. Kim et al.~\cite{Knowledge-Grounded-KE-T5}, PARC~\cite{PARC}, and CREA-ICL~\cite{CREA-ICL} improve the caliber of non-English conversations by incorporating cross-lingual knowledge, effectively addressing the scarcity of non-English datasets and enhancing the quality of the generated dialogue.
CEG~\cite{CEG} addresses hallucination issues through a post-processing mechanism, verifying LLM-generated answers through retrieval.

\subsubsection{\textbf{Neural Machine Translation}}

Neural Machine Translation (NMT) is the automated process of translating text from a source language to a target language~\cite{TRIME,NMT-with-Monolingual-TM,KNN-MT}. It is a pivotal task in the domain of NLP and represents a significant objective in the pursuit of AI, boasting considerable scientific and practical significance.

Cai et al.~\cite{NMT-with-Monolingual-TM} proposed an innovative approach that utilizes monolingual corpora alongside multilingual learning techniques, challenging the traditional dependency on bilingual corpora in Neural Machine Translation. 
kNN-MT~\cite{KNN-MT} executes translation tasks at the token level by computing vector space distances. 
TRIME~\cite{TRIME} effectively minimizes the discrepancy between training and inference phases by jointly training the retrieval system and the generation model, thereby enhancing the precision of translations.
\subsubsection{\textbf{Event Extraction}}

Event extraction is a process in NLP that involves identifying and categorizing specific events within a text and associating them with relevant entities. These events are usually represented by verbs and the entities are the participants involved in the event.
R-GQA~\cite{R-GQA} enhances the context of a given issue by identifying and utilizing the most closely aligned Question-Answer pair from a repository, thereby enriching the information available for processing the current query.

\subsubsection{\textbf{Summarization}}
Summarization is a task aimed at distilling the essential information from lengthy texts and producing a concise, coherent summary that encapsulates the primary themes. There are two main approaches to summarization: extractive and abstractive.

Extractive summarization involves the automatic selection and compilation of key phrases directly from the source text, which refrains from creating new sentences, instead repurposing segments from the original text.

Abstractive summarization, on the other hand, entails comprehending the original text's meaning and reformulating it into new sentences~\cite{RAMKG,Unlimiformer,RPRR,RIGHT}, which can convey the source's intent more fluidly but poses greater challenges in terms of implementation due to its complexity.
RAMKG~\cite{RAMKG} effectively leverages a comprehensive English corpus to bolster the performance of keyphrase generation in non-English contexts.
Unlimiformer~\cite{Unlimiformer} addresses the issue of input length constraints in transformer-based models by retrieving and utilizing the top-k most relevant hidden states, thereby extending the model's capacity to handle longer inputs. 
RPRR~\cite{RPRR} employs a Retrieve-Plan-Retrieve-Read approach to overcome the limited context window constraints faced by LLMs, utilizing retrieved information to generate high-quality Wikipedia documents for emerging events. 
RIGHT~\cite{RIGHT} chooses to use different types of retrievers in different datasets to enhance the generator.
M-RAG~\cite{M-RAG} significantly enhances text summarization by segmenting documents into various databases and incorporating multi-agent reinforcement learning techniques.

\subsection{\textbf{RAG for Code}}

Separate retrieval and generation approaches have historically been employed for code-related tasks.
For retrieval, similar code snippets can be identified using Abstract Syntax Trees (AST) or text edit distance.
%
For generation, sequence-to-sequence models are employed to generate code or natural language. 
Recent RAG research combines both retrieval and generation techniques to enhance the overall performance.

\subsubsection{\textbf{Code Generation}}

Code generation aims to convert Natural Language (NL) descriptions into code implementations. 

Query-based RAG is a common method for code generation.
It builds prompts for transformer-based generative models with retrieved information, including similar examples~\cite{DBLP:conf/emnlp/ParvezACRC21,wang2022codet5mix,DBLP:conf/emnlp/MadaanZ0YN22,DBLP:conf/emnlp/WangLGB0H23,li2023acecoder,chen2024code}, relevant API details~\cite{DBLP:conf/emnlp/ZanCLGWL22,zan2023private}, documentations~\cite{DBLP:conf/iclr/Zhou0XJN23}, imports~\cite{DBLP:conf/kbse/LiuYLDWP23}, and global functions~\cite{liao2023context}.
SKCODER~\cite{li2023skcoder} retrieves relevant code snippets to produce sketch template for final code generation.
RRGCode~\cite{gou2024rrgcode} employs a cross-encoder to rank the retrieval results.
CODEAGENT~\cite{DBLP:journals/corr/abs-2401-07339} designs agents for web search, documentation retrieval, program generation, and correctness testing.
ARKS~\cite{su2024arks} incorporates iterative RAG to re-formulate queries and update retrieval sources.

Logit-based RAG is also applicable for code generation.
RECODE~\cite{DBLP:conf/emnlp/HayatiOAYTN18} retrieves NL descriptions and paired codes using edit distance, then extracts n-gram action subtrees from ASTs. During LSTM-based generation, the processed subtrees are leveraged through logits at each decoding step. 
kNN-TRANX~\cite{DBLP:conf/emnlp/Zhang0YC23} uses a seq2tree model to convert NL to code AST. 
%
During each decoding step, hidden states are searched in the AST prefix datastore to create new probabilities, later merged with the seq2tree model's output via a confidence network.

ToolCoder~\cite{DBLP:journals/corr/abs-2305-04032} generates codes containing special tokens. When it encounters these tokens, ToolCoder performs online search or offline retrievals to fill in the blanks with API calls, which is a specialized form of speculative RAG.

\subsubsection{\textbf{Code Summarization}}

Code summarization tasks in turn convert the code into NL descriptions.

Many research works process retrieval results using additional encoders and then combine them for subsequent decoder, which is similar to the Fusion-in-Decoder~\cite{FId}.

\begin{figure}[h]
  \centering
  \includegraphics[width=\linewidth]{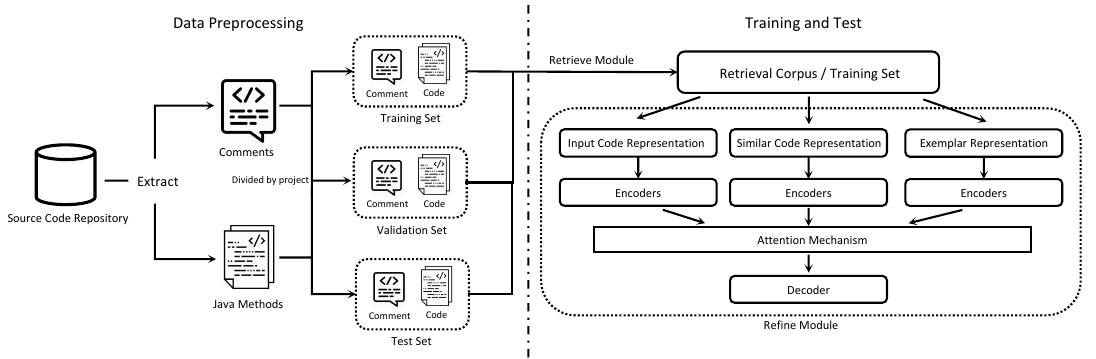}
  \caption{Architecture of Re2Com~\cite{DBLP:conf/kbse/WeiLLXJ20} model.}
  \label{fig:re2com}
\end{figure}
Re2Com~\cite{DBLP:conf/kbse/WeiLLXJ20} and EditSum~\cite{DBLP:conf/kbse/LiL000J21} retrieve similar codes using BM25 and generate summary using LSTM. 
They separately encode the input, the retrieved code, and the corresponding summary, then combine the hidden states or logits in the decoder. 
HGNN~\cite{DBLP:conf/iclr/LiuCXS021} instead uses code edit distance for retrieval, and substitutes the code encoder with hybrid GNN on their Code Property Graphs (CPG)~\cite{DBLP:conf/sp/YamaguchiGAR14}. 
RACE~\cite{DBLP:conf/emnlp/ShiW0DZHZ022} employs separate encoders for the input code difference, the retrieved code differences through dense retrieval, and corresponding commit message to generate the final commit messages. 
BASHEXPLAINER~\cite{DBLP:conf/icsm/YuYCLZ22} applies dense retrieval, and fuses the embeddings for subsequent transformer-based decoder. 
READSUM~\cite{choi2023readsum} uses Levenshtein distance for retrieval, and employs a fusion network to combine the representations of retrieved codes and summaries.

Query-based RAG is prevalent for code summary generation. 
REDCODER~\cite{DBLP:conf/emnlp/ParvezACRC21}, ASAP~\cite{ahmed2024automatic}, and SCCLLM~\cite{DBLP:journals/infsof/Zhao0YS24} all form prompts with retrieved contents for summarization. They employ dense retrieval, sparse retrieval, and hybrid retrieval (including semantic, syntactic, and lexical-based retrieval), respectively. 
The paradigm is also leveraged for pseudocode generation~\cite{alokla2022retrieval} and log statement generation~\cite{DBLP:conf/icse/XuCZZHHL0LDRZ24}. 

Logit-based RAG also prevails in code summarization.
Rencos~\cite{DBLP:conf/icse/ZhangW00020} and CoRec~\cite{DBLP:journals/tosem/WangXLHWG21} retrieve similar code snippets or code differences through AST or dense representations. They both adopt multiple LSTMs for the input and the retrieved results, and the probabilities are combined for final generation. 
kNN-Transformer~\cite{zhu2022simple} uses a transformer-based generator to obtain context vectors of input codes, then combines three parts of logits from vector search, the generator, and the copy mechanism for rare tokens in the input.
Tram~\cite{DBLP:journals/corr/abs-2305-11074} also combines three sets of logits from the original generator, the generator for sentence-level retrieved results, and the search logits of the token-level vectors (which represent the source codes and their ASTs).
CMR-Sum~\cite{li4724884cross} incorporates the cross-attention probabilities between the retrieved summary and the generated summary, to the original generation logits.

\subsubsection{\textbf{Code Completion}}

Code completion is akin to the code version of the “next sentence prediction” task. 

Query-based RAG is the mainstream paradigm for code completion.
Drain et al.~\cite{drain2021generating} retrieved template functions for function completion.
ReACC~\cite{DBLP:conf/acl/LuDHGHS22} uses both sparse and dense retrieval. 
RepoCoder~\cite{DBLP:conf/emnlp/ZhangCZKLZMLC23} performs iterative RAG by augmenting the retrieval input with previously generated code.
De-Hallucinator~\cite{DBLP:journals/corr/abs-2401-01701} retrieves API references using first-time generated contents, then conducts query-based RAG for improved code completion. 
REPOFUSE~\cite{liang2024repofuse} includes rationale context and retrieved codes to form prompt, and ranks the contexts to fit in the length limit. 

Many works leverage latent representation-based RAG.
Retrieve-and-edit~\cite{DBLP:conf/nips/HashimotoGOL18}, RepoFusion~\cite{shrivastava2023repofusion}, and EDITAS~\cite{DBLP:conf/kbse/SunLYLZ23} employ multiple encoders for retrieved contents or edit sequences, then fuse the information for subsequent decoder.
CoCoMic~\cite{DBLP:journals/corr/abs-2212-10007} retrieves codes on the project context graph of the whole code project. It jointly processes the representations of source codes and retrieved contexts in the generator.

kNM-LM~\cite{tang2023domain} performs logit-based RAG, combining the logits of retrieval and generation using bayes inference.
%

\subsubsection{\textbf{Automatic Program Repair}}

Query-based RAG is often used in automatic program repair to help generative models fix buggy codes. 
RING~\cite{DBLP:conf/aaai/JoshiSG0VR23}, CEDAR~\cite{DBLP:conf/icse/NashidSM23}, and RAP-Gen~\cite{DBLP:conf/sigsoft/Wang0JH23} all use hybrid retrieval (including both sparse and dense retrieval) for similar error messages, buggy codes, or fixes to build prompts.
InferFix~\cite{DBLP:conf/sigsoft/JinSTSLSS23} includes the bug type, the location, relevant syntax hierarchies, and similar fixes into the prompt.
SARGAM~\cite{DBLP:journals/corr/abs-2306-06490} utilizes prompts with similar buggy codes to generate patches; then another model is employed to refine the final result.
RTLFixer~\cite{DBLP:journals/corr/abs-2311-16543} leverages ReAct~\cite{ReAct} to implement an agent fixing errors in Verilog codes. It iteratively retrieves errors and paired solutions, and combines reasoning and action planning into prompts for LLMs. 

\subsubsection{\textbf{Text-to-SQL and Code-based Semantic Parsing}}

Semantic parsing converts NL into clear, structured representations, like SQL or other domain-specific languages, often with the assistance of codes. 
All related works that employ RAG specifically utilize its query-based variant.
XRICL~\cite{DBLP:conf/emnlp/0010Z0L22} searches and reranks English utterance using non-English ones, then builds prompt to generate SQL queries. 
SYNCHROMESH~\cite{DBLP:conf/iclr/PoesiaP00SMG22} retrieves similar NL and SQL to build prompts, then conducts constrained semantic decoding to enforce rich syntactic and semantic constraints during SQL generation.
CodeICL~\cite{bogin2023leveraging} uses Python for semantic parsing, leveraging BM25 to incorporate similar training examples into prompts.
RESDSQL~\cite{li2023resdsql} includes ranked schemas into prompts to generate SQL skeleton and SQL query.
ReFSQL~\cite{DBLP:conf/emnlp/ZhangLWZSCTJS23} uses a structure-enhanced retriever with schema linking and Mahalanobis contrastive learning, which helps to make better text-to-SQL generation.
To build prompts for SQL generation, ODIS~\cite{chang2023selective} retrieves both in-domain and out-of-domain demonstrations, while Nan et al.~\cite{nan2023enhancing} retrieved both similar and diverse demonstrations.
MURRE~\cite{zhang2024multi} conducts multi-hop retrieve-rewrite on tables to generate tabularized question, then ranks the results for prompt construction.
CodeS~\cite{li2024codes} retrieves relevant information from table databases in a coarse-to-fine manner to generate SQL.

\subsubsection{\textbf{Others}}

There are several other code-related tasks that adopt query-based RAG paradigm, incorporating similar examples to construct prompts.
Jie et al.~\cite{jie2023leveraging} used programs as the intermediate step in numerical reasoning.
De-fine~\cite{gao2023fine} uses programs to solve complex tasks. It refines the answer generated by query-based RAG, then adds the refined programs back to the retrieval source.
For program static analysis, E\&V~\cite{hao2023v} leverages an LLM agent to form intermediate results with AST-based source code retrieval, pseudo-code execution, execution specifications verification, and other tools.
Code4UIE~\cite{DBLP:journals/corr/abs-2311-02962} performs information extraction through code representation. 
StackSpotAI~\cite{pinto2024lessons} builds an AI coding assistant with an RAG component.
InputBlaster~\cite{DBLP:journals/corr/abs-2310-15657} generates unusual text input that could cause mobile app crash.

\subsection{\textbf{RAG for Knowledge}}

Structured knowledge, including KGs (Knowledge Graph) and tables, is widely used in language-related tasks. It usually serves as the retrieval source to augment generation.
In addition to regular sparse and dense retrieval, NER (Named-Entity Recognition) technique and graph-aware neighbor retrieval are applied to identify and extract relevant entities and relations.
\subsubsection{\textbf{Knowledge Base Question Answering}}

KBQA (knowledge base question answering) typically utilizes a knowledge base to determine the correct answer to a question. Many semantic parsing methods have been proposed, generating logical forms (e.g. SPARQL) based on the question.

Query-based RAG is the mainstream approach. 
Unseen Entity Handling~\cite{DBLP:conf/emnlp/HuangKZ21} uses FreeBase~\cite{DBLP:conf/sigmod/BollackerEPST08} to retrieve topic entities, which are combined with query to generate SPARQL output. 
CBR-KBQA~\cite{DBLP:conf/emnlp/DasZTGPLTPM21} combines the query and the retrieved (query, logical form) pairs for generation. It also revises the final result to align with the relations present in the knowledge graph.
GMT-KBQA~\cite{DBLP:conf/coling/HuWSQ22} re-ranks the retrieved entities and relations, and conducts relation classification and entity disambiguation before generation.
RNG-KBQA~\cite{DBLP:conf/acl/YeYHZX22}, TIARA~\cite{DBLP:journals/corr/abs-2210-12925}, BLLM augmentation~\cite{DBLP:journals/corr/abs-2311-08894}, and Shu et al.~\cite{DBLP:journals/corr/abs-2309-08345} re-rank the candidate logical forms or entities from the knowledge graph for prompt construction.
Uni-Parser~\cite{DBLP:conf/emnlp/LiuYMRXZ22} includes entities from mention detection, 2-hop paths extraction, and tables from databases into generator input.
ECBRF~\cite{DBLP:conf/eacl/YangDCC23} follows the case-based reasoning paradigm~\cite{DBLP:conf/iccbr/LeakeC20}, retrieving similar triplet to build prompt input.
FC-KBQA~\cite{DBLP:conf/acl/ZhangZWCHL0L23} extracts relevant classes, relations, and entities from BM25 or mention detection, StructGPT~\cite{DBLP:conf/emnlp/JiangZDYZW23} extracts relevant triplets and nearest entities, and KAPING~\cite{DBLP:journals/corr/abs-2306-04136} extracts relevant facts through entity matching.
Sen et al.~\cite{sen2023knowledge} replaced the retrieval with a relation distribution generation model for weighted triplets.
Retrieve-Rewrite-Answer~\cite{DBLP:journals/corr/abs-2309-11206} retrieves subgraphs into prompts using hop prediction, relation path prediction, and triplet sampling.
Keqing~\cite{DBLP:journals/corr/abs-2401-00426} decomposes a complex question into simple sub-questions through LLM, then retrieves sub-question templates and extract candidate entities from knowledge graph, and finally generates the answer through ChatGPT.
Liu et al.~\cite{DBLP:journals/corr/abs-2401-05777} leveraged retrieved pairs to explore the capability of formal language understanding and generation.
Interactive-KBQA~\cite{xiong2024interactive} employs the LLM as an agent, which conducts entity-linking on KG and generates current thought and action until obtaining the final answer.

Latent representation-based RAG is also employed for KBQA.
ReTraCk~\cite{DBLP:conf/acl/ChenLYLLJ21} retrieves entities and schemas through mention detection and dense retrieval. It generates logical forms using LSTM, using retrieved items through knowledge-specific rules.
SKP~\cite{DBLP:conf/cikm/DongLWZXX23}, DECAF~\cite{DBLP:conf/iclr/YuZNZL0HWWX23}, and KD-CoT~\cite{DBLP:journals/corr/abs-2308-13259} all retrieve triplets and conduct fusion-in-decoder~\cite{FId} RAG. 
KD-CoT also follows a chain-of-thought paradigm, iteratively performing retrieval, generation, and verification. 

\subsubsection{\textbf{Knowledge-augmented Open-domain Question Answering}}

Structured knowledge is often leveraged to augment ODQA (open-domain question answering).

Latent representation-based RAG, especially the fusion-in-decoder~\cite{FId} technique, is prevalent for knowledge-augmented ODQA.
UniK-QA~\cite{DBLP:conf/naacl/OguzCKPOSGMY22}, KG-FiD~\cite{DBLP:conf/acl/Yu0F0WXRY022}, GRAPE~\cite{DBLP:conf/emnlp/Ju00Z022} all apply the fusion-in-decoder technique. They incorporate triplet-based documents, re-ranked documents through KG, and bipartite graph for pairs of question and passage, respectively.
\begin{figure}[h]
  \centering
  \includegraphics[width=\linewidth]{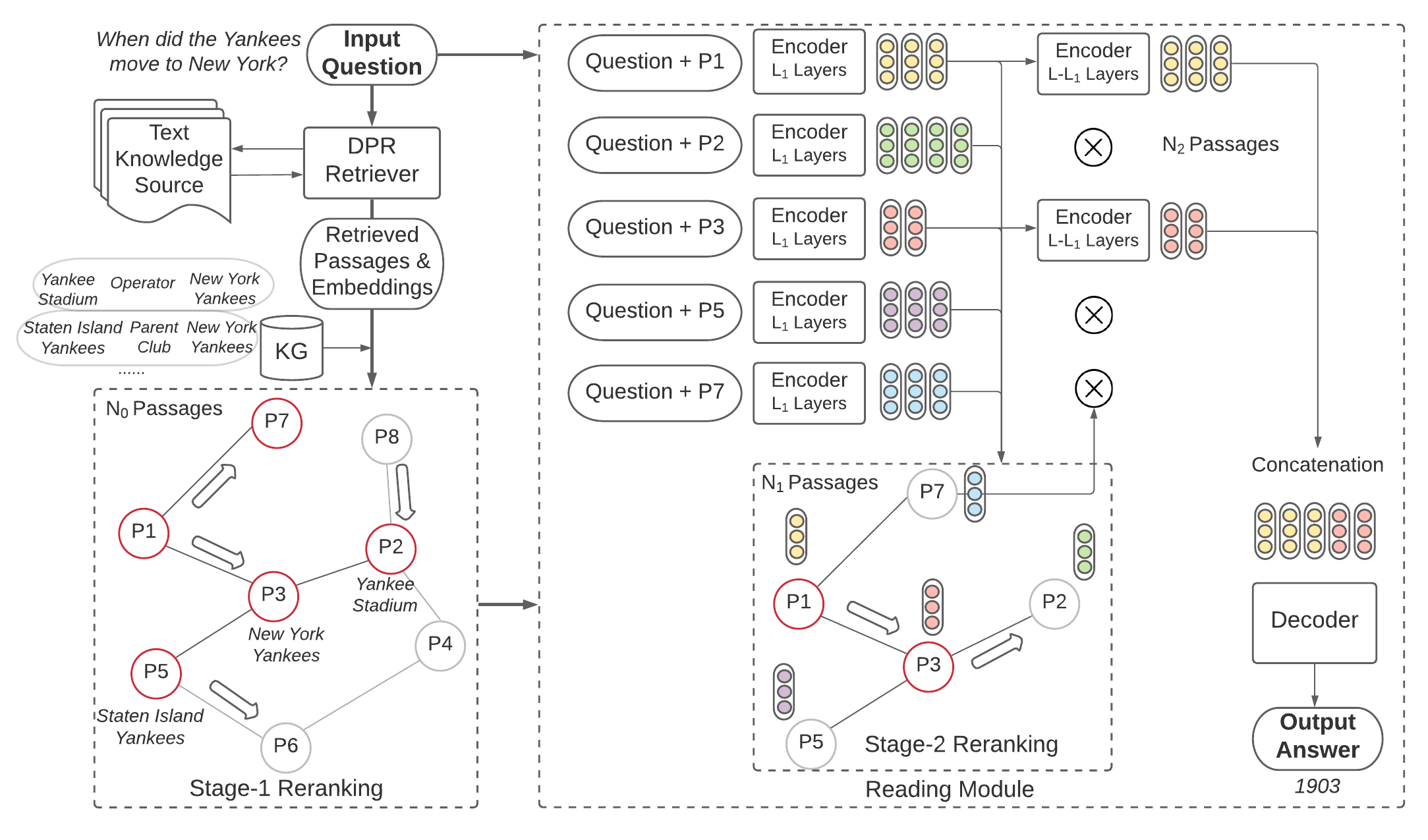}
  \caption{Architecture of KG-FiD~\cite{DBLP:conf/acl/Yu0F0WXRY022} model.}
  \label{fig:kg-fid}
\end{figure}
OREOLM~\cite{DBLP:conf/emnlp/HuX0WYZCS22} empowers LLM with knowledge reasoning paths, integrating the entity value memory derived from contextualized random walk paths on KG into the hidden states of the LLM.
SKURG~\cite{DBLP:conf/mm/YangCWHZ23} performs iterative retrieval and generation, using cross-attention to incorporate data sources into the input embedding. It uses a gate score to determine whether to re-start retrieval or to generate the real answer.

With the rapid development of LLMs, query-based RAG is emerging as a new standard.
DIVKNOWQA~\cite{DBLP:journals/corr/abs-2310-20170} retrieves from multiple sources using different techniques. It iteratively retrieves and re-ranks the data before generating the final answer. 
KnowledGPT~\cite{DBLP:journals/corr/abs-2308-11761} uses generated code to retrieve from both public and personal knowledge bases.
EFSUM~\cite{ko2024evidence} optimizes the evidence-focused summary after facts-augmented generation, so as to align the QA-specific preference for helpfulness and faithfulness.
GenTKGQA~\cite{gao2024two} employs GNN (graph neural network) to integrate structural and temporal information from subgraph retrieval into virtual token representations.
KnowledgeNavigator~\cite{guo2023knowledgenavigator} performs retrieval on KG through iterative filtering of relations with respect to core entities, so as to obtain relevant triplets.

GNN-RAG~\cite{GNN-RAG} fuses LLMs' language understanding with GNN's reasoning prowess and employs a retrieval augmentation strategy to enhance KGQA performance.
\subsubsection{\textbf{Table for Question Answering}}
Tables, as another form of structured knowledge, also facilitates question answering.

Fusion-in-decoder~\cite{FId} style RAG is often used for table QA.
EfficientQA~\cite{min2021neurips}, a competition held in NeurIPS 2020, witnessed the proposal of numerous retrieval-reader systems that rely on textual and tabular data.
Dual Reader-Parser~\cite{DBLP:conf/acl/LiNXZWX20} and CORE~\cite{DBLP:conf/emnlp/Ma00NG22} both re-rank the retrieved textual and tabular data for generation.
Convinse~\cite{DBLP:conf/sigir/ChristmannRW22} retrieves information from knowledge bases, tables, and texts after question understanding.
RINK~\cite{park2023rink} designs a set-level reader-inherited re-ranker to get the relevance score of table segments.
TAG-QA~\cite{DBLP:journals/corr/abs-2309-11049} retrieves tables and texts through GNN (after table-to-graph conversion) and BM25, respectively.

Tables can be integrated into prompts for query-based RAG.
Both T-RAG~\cite{pan2022end} and OmniTab~\cite{DBLP:conf/naacl/JiangMHNC22} concatenates the retrieved tables with the query to generate the answer.
CARP~\cite{DBLP:conf/ijcai/ZhongHL0W0D22} extracts hybrid chain of retrieved tables and passages for prompt construction.
StructGPT~\cite{DBLP:conf/emnlp/JiangZDYZW23} retrieves from multiple sources including KGs, tables, and databases.
%
cTBLS~\cite{sundar2023ctbl} forms prompts with ranked tables after retrieval.
Min et al.~\cite{min2024exploring} integrated tabular data through table-to-text techniques, then experiments on both finetuning and RAG.
ERATTA~\cite{ERATTA} generates SQL code to extract table information, integrating it into the prompt to minimize model hallucination.

\subsubsection{\textbf{Others}}

Prototype-KRG~\cite{wu2020improving} integrates retrieved knowledge facts and dialogue prototypes into a GRU model through both hidden states and logits.
SURGE~\cite{kang2022knowledge} combines relevant subgraphs into the input for dialogue generation.
RHO~\cite{DBLP:conf/acl/JiLLYWZF23} fuses KG embedding of relevant entities and relations into textual embeddings during dialogue generation.
K-LaMP~\cite{DBLP:journals/corr/abs-2311-06318} retrieves entities in history queries to construct prompt for query suggestion.
ReSKGC~\cite{DBLP:conf/sigir/YuY23} retrieves relevant triplets to complete triplet using Fid.
G-Retriever~\cite{DBLP:journals/corr/abs-2402-07630} retrieves nodes and edges from textual graphs to construct subgraph and perform graph prompt tuning for QA.
Hussien et al.~\cite{hussien2024rag} fuse the reasoning power of KG with the expressiveness of LLMs through RAG techniques.
HippoRAG~\cite{HippoRAG} excels in multi-hop question answering by emulating mammalian brain knowledge storage with KG triples and employing a personalized PageRank algorithm for retrieval.
\subsection{\textbf{RAG for Image}}
\subsubsection{\textbf{Image Generation}}
Image generation refers to the process of creating new images, typically using algorithms in the field of artificial intelligence and machine learning.

The retrieval process can not only help yield high-quality images even for rare or unseen subjects, but also reduces the parameter count and computational expense \cite{tseng2020retrievegan,casanova2021instance,sheynin2022knn,blattmann2022retrieval,chen2022re,kirstain2023x,rombach2022text,li2022memory}.
For GAN-based model, 
RetrieveGAN \cite{tseng2020retrievegan} uses a differentiable retriever for image patch selection, facilitating end-to-end training. IC-GAN \cite{casanova2021instance} models data as conditional distributions around each training instance, conditioning both the generator and discriminator on these instances.

Recently, diffusion models beat GANs on image generation \cite{dhariwal2021diffusion}.
KNN-Diffusion \cite{sheynin2022knn} and RDM \cite{blattmann2022retrieval} train diffusion models conditioned on CLIP embeddings and image neighbors, enabling post-hoc conditioning on labels, prompts, and zero-shot stylization \cite{rombach2022text}.
Beyond only images, Re-imagen \cite{chen2022re} extends retrieval to image-text pairs for text-to-image generation, with interleaved guidance to balance the alignment between prompts and retrieval conditions.
Retrieve\&Fuse \cite{kirstain2023x} prevents information loss of CLIP embeddings by concatenating retrieved and noised images before each U-Net attention block, allowing fully interaction via self-attention. 
RPG \cite{yang2024mastering} retrieves representative images to construct in-context examples, and utilizes chain-of-thought reasoning \cite{zhang2023multimodal} to plan out complementary subregions for compositional text-to-image diffusion.

\subsubsection{\textbf{Image Captioning}}
Image captioning is the process of generating a textual description of an image.

Retrieval-augmented image captioning typically synthesises description with a collection of retrieved captions.
MA~\cite{fei2021memory} augments via a memory bank, built with historical context and target word of image-text training set, and queried with inference context.
In adversarial training, RAMP \cite{xu2021retrieval} takes retrieved captions as discriminator reference, and employs memory-augmented attention and copying mechanisms for better utilization of retrieved captions.
\begin{figure}[h]
  \centering
  \includegraphics[width=\linewidth]{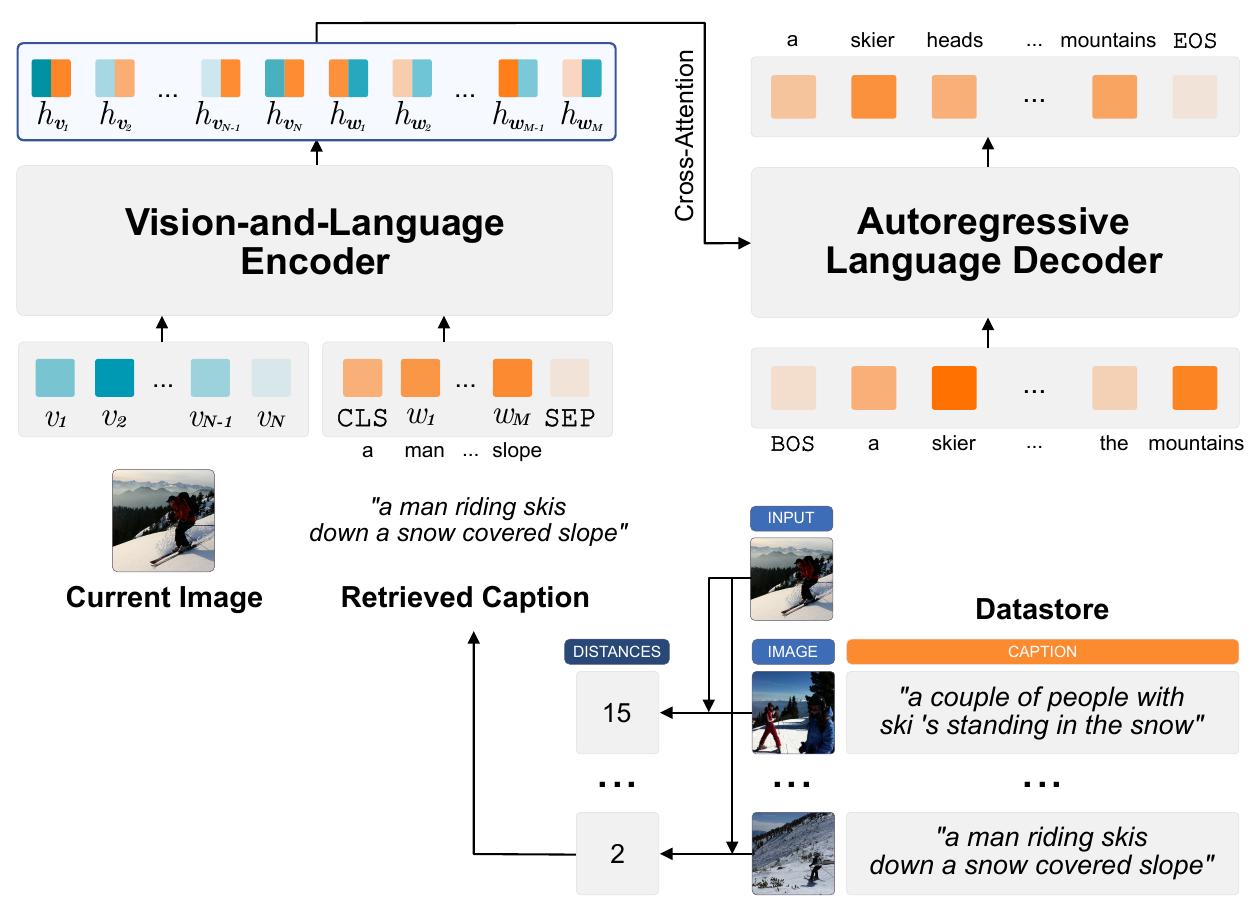}
  \caption{Architecture of EXTRA~\cite{ramos2023retrieval} model.}
  \label{fig:extra}
\end{figure}
The RA-Transformer \cite{sarto2022retrieval} and EXTRA \cite{ramos2023retrieval}, both retrieval-augmented transformer-based captioning models, utilize cross-attention over encoded retrieved captions.
Beyond caption retrieval, REVEAL \cite{hu2023reveal} uniformly encodes and retrieves multi-modal world knowledge, integrated with retrieval score-aware attention.
Directly, SMALLCAP \cite{ramos2023smallcap} employs a CLIP vision encoder and a LLM decoder, with retrieved captions serving as input-specific in-context examples.
For remote sensing images, CRSR \cite{li2024cross} refines retrieved captions, filtering out misleading details and emphasizing visually salient content.

\subsubsection{\textbf{Others}}
There also exist many retrieval augmented works for other image-related tasks. For Visual Question Answering (VQA), PICa \cite{yang2022empirical} converts images into textual descriptions, prompts GPT-3 and ensembles multi-query results. RA-VQA \cite{lin2022retrieval} enables an end-to-end training with differentiable retrieval for answer generation. For visually grounded dialogue, KIF \cite{fan2021augmenting} and Maria \cite{liang2021maria} enhances dialog generation with external knowledge like visual experiences. In multi-modal machine translation, \cite{fang2022neural} incorporates visual information at the phrase level to improve NMT with multi-modal information.

\subsection{\textbf{RAG for Video}}
\subsubsection{\textbf{Video Captioning}}
Video captioning translates the visual content into descriptive utterances.
KaVD~\cite{whitehead2018incorporating} generates news video caption with background knowledge in related documents like named entities and events.
R-ConvED~\cite{DBLP:journals/tomccap/ChenPLYCM23} retrieves relevant sentences and videos via Dual Encoding~\cite{DBLP:conf/cvpr/DongLXJH0W19}, and predicts the target word with a convolutional encoder-decoder network.
CARE~\cite{DBLP:journals/tip/YangCZ23} combines three modalities data, i.e. frame, audio, and retrieved texts, to provide both global and local semantic guidance as augmentation.
EgoInstructor~\cite{DBLP:journals/corr/abs-2401-00789}  focuses on first-person videos, retrieves relevant exocentric videos and texts, and generates captions through LLM via cross-attention with encoded videos

\subsubsection{\textbf{Video QA\&Dialogue}}
Video QA\&Dialogue generates single or multiple-round responses in alignment with video content. 
For VideoQA, MA-DRNN \cite{yin2019memory} stores and retrieves useful information in queries and videos with external memory, therefore models the long-term visual-textual dependence. R2A \cite{pan2023retrieving} retrieves semantically similar texts by CLIP, and prompts LLM with both the query and the retrieved texts. For video dialogue, \cite{lei2019tvqa+} proposes TVQA+ dataset to enable relevant moments and visual concepts retrieval, and designs corresponding spatio-temporal-aware generator. VGNMN \cite{le2022vgnmn} extracts visual cues from videos, while the retrieval process is parameterized by entities and actions in previous dialogues. 

\subsubsection{\textbf{Others}}
RAG also works for other video-related tasks. 
VidIL \cite{wang2022language} converts video content into temporal-aware LLM prompts for tasks like video captioning, question answering, and future event prediction. 
\begin{figure}[h]
  \centering
  \includegraphics[width=\linewidth]{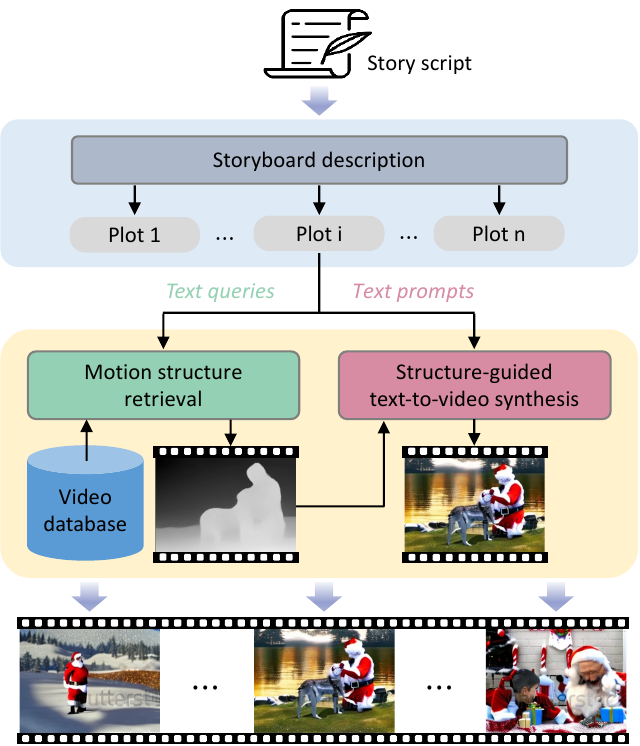}
  \caption{Architecture of Animate-A-Story~\cite{DBLP:journals/corr/abs-2307-06940} model.}
  \label{fig:animate-a-story}
\end{figure}
For trustworthy autonomous driving, RAG-Driver \cite{yuan2024rag} grounds the MLLM in retrieved expert demonstrations, to produce driving action explanations. 
Animate-A-Story~\cite{DBLP:journals/corr/abs-2307-06940} simplifies text-to-video generation by dividing it into plot-based video augmentation and video-diffusion generation conditioned on text and video inputs.

\subsection{\textbf{RAG for Audio}}
\subsubsection{\textbf{Audio Generation}}

Audio generation usually synthesises audio with natural language prompt.  
\begin{figure}[h]
  \centering
  \includegraphics[width=\linewidth]{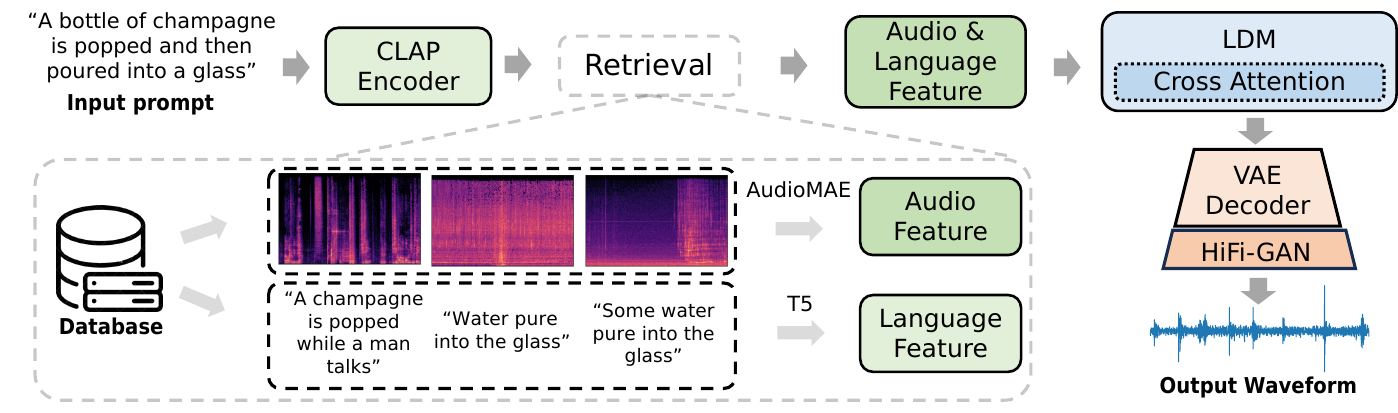}
  \caption{Architecture of Re-AudioLDM~\cite{DBLP:journals/corr/abs-2309-08051} model.}
  \label{fig:audioldm}
\end{figure}
Given input prompt, Re-AudioLDM~\cite{DBLP:journals/corr/abs-2309-08051} retrieves relevant caption-audio pairs with dense retriever CLAP~\cite{DBLP:conf/icassp/WuCZHBD23} for generation.
Make-An-Audio~\cite{DBLP:conf/icml/HuangHY0LLYLYZ23} retrieves audios given text prompt, then constructs pseudo prompts for text-to-audio diffusion model training.

\subsubsection{\textbf{Audio Captioning}}
Audio captioning, basically a sequence-to-sequence task, generates natural language data for audio data. 
RECAP~\cite{DBLP:journals/corr/abs-2309-09836} and~\cite{DBLP:journals/corr/abs-2012-07331} leverages dense retrievers, CLAP \cite{DBLP:conf/icassp/WuCZHBD23} and VGGish~\cite{DBLP:conf/icassp/HersheyCEGJMPPS17} respectively, to retrieve related captions given audio data. For RECAP, captions are included into LLM prompts, while ~\cite{DBLP:journals/corr/abs-2012-07331} uses both audio and retrieved captions in attention module.
Other research studies align audio modality with text to leverage advancements in LLMs~\cite{DBLP:journals/corr/abs-2309-05767,DBLP:journals/corr/abs-2309-12242,DBLP:journals/corr/abs-2309-07372} for various downstream text generation.

\subsection{\textbf{RAG for 3D}}
\subsubsection{\textbf{Text-to-3D}}
Retrieval can be applied to augment 3D asset generation.

\begin{figure}[h]
  \centering
  \includegraphics[width=\linewidth]{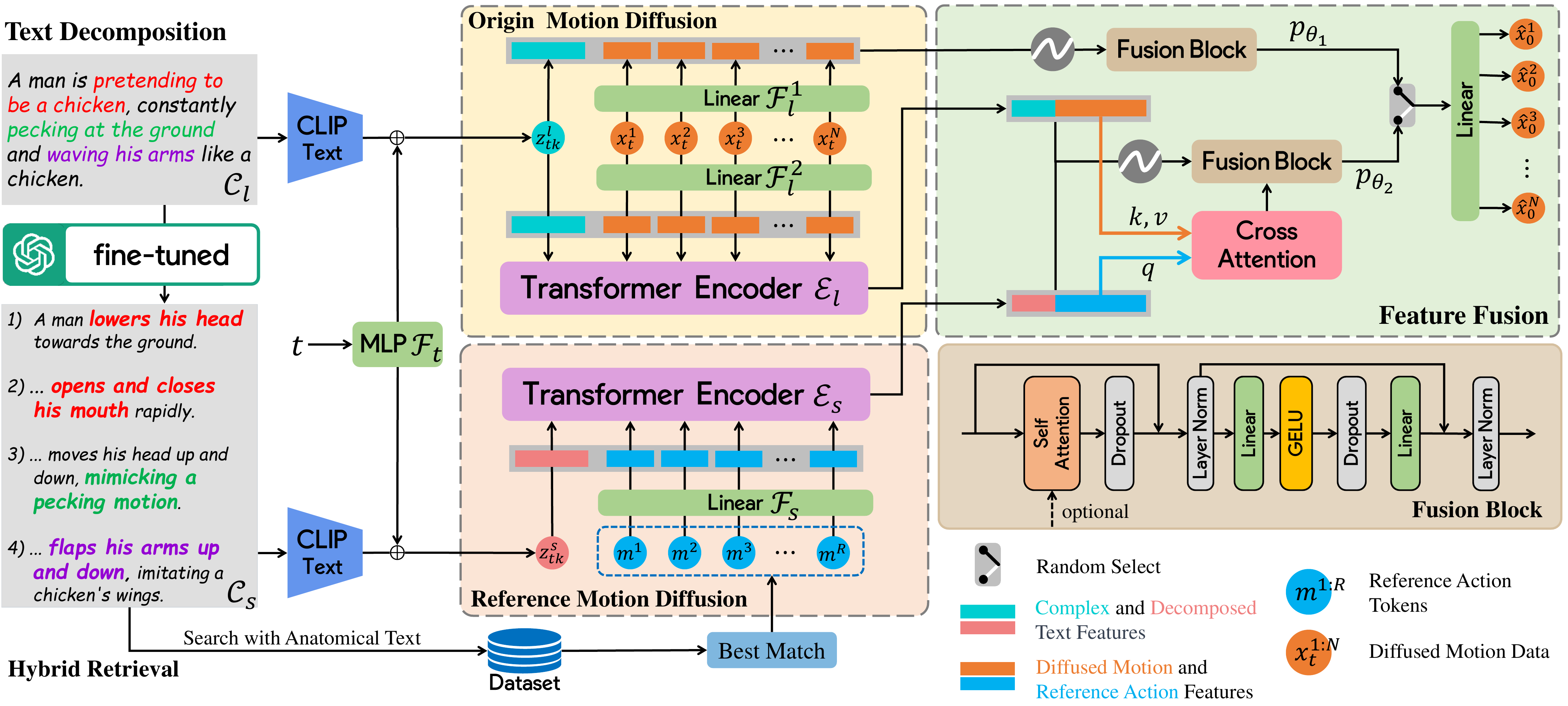}
  \caption{Architecture of AMD~\cite{jing2023amd} model.}
  \label{fig:amd}
\end{figure}

ReMoDiffuse~\cite{DBLP:conf/iccv/ZhangGPCHLYL23} retrieves relevant motion entities and generates motions using diffusion models, with the semantic-modulated attention and condition mixture guidance.
AMD~\cite{jing2023amd} designs and fuses two motion diffusion models. One branch conditions on the original prompt, while the other decomposes the prompt into anatomical scripts and retrieves similar motions.
RetDream~\cite{DBLP:journals/corr/abs-2402-02972} retrieves 3D assets to augment the variational score distillation \cite{wang2024prolificdreamer} of 2D diffusion models. These assets offer geometric and adapted 2D priors, which not only impose additional velocity on particles for initialization but also help optimize 2D diffusion models by LoRA.

\subsection{\textbf{RAG for Science}}

RAG has also emerged as a promising research direction for many interdisciplinary applications, such as molecular generation, medical tasks and computational research.

\subsubsection{\textbf{Drug Discovery}}
The goal of drug discovery is to generate molecules that concurrently fulfill diverse properties.

\begin{figure}[h]
  \centering
  \includegraphics[width=\linewidth]{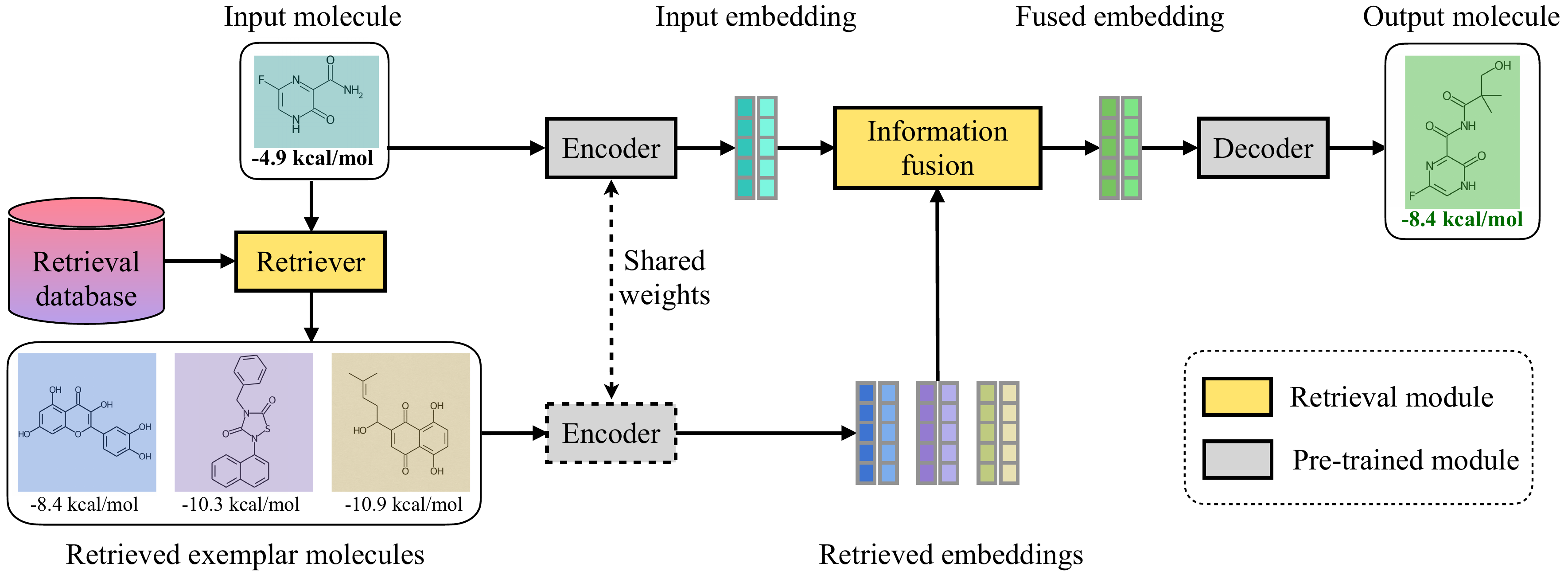}
  \caption{Architecture of RetMol~\cite{wang2022retrieval} model.}
  \label{fig:retmol}
\end{figure}

RetMol~\cite{wang2022retrieval} integrates a lightweight retrieval mechanism and molecular strings into a pre-trained encoder-decoder generative model to retrieve and fuse exemplar molecules with the input.
PromptDiff~\cite{yang2023prompt} introduces an interaction-based, retrieval-augmented 3D molecular diffusion model that retrieves a curated set of ligand references to guide the synthesis of ligands meeting specific design criteria.

\subsubsection{\textbf{Biomedical Informatics Enhancement}}
Several recent studies have improved the expressiveness of LLM by retrieving information from biomedical domain-specific databases, thereby augmenting the model's capabilities to provide valuable guidance for tasks in the medical field.

PoET~\cite{truong2024poet} is an autoregressive model using a transformer variant with a retrieval mechanism for prompt augmentation, speeding up the prediction of protein variant fitness properties.
Chat-Orthopedist~\cite{shi2023retrieval} enhances ChatGPT with a retrieval-augmented mechanism focused on adolescent idiopathic scoliosis (AIS), utilizing an external knowledge base for precise responses.
BIOREADER~\cite{frisoni2022bioreader} is the first retrieval-enhanced text-to-text transformer-based model for biomedical natural language processing, incorporating the retrieved literature evidence into the model using a chunked-cross attention mechanism.
MedWriter~\cite{yang2021writing} employs a hierarchical retrieval-augmented generation method that combines report-level and sentence-level templates to produce coherent and clinically accurate medical reports from images.
QA-RAG~\cite{kim2024rag} employs a dual-track RAG strategy to enhance pharmaceutical compliance by effectively retrieving and integrating regulatory guidelines based on language model responses and user queries.\
RAG-RLRC-LaySum~\cite{RAG-RLRC-LaySum} leverages biomedical text knowledge for llms, employing reinforcement learning and re-ranking techniques to enhance content relevance and readability of the output.
\subsubsection{\textbf{Math Applications}}
Retrieval-augmented generation technology in mathematics streamlines problem-solving, boosts research innovation, and refines educational strategies.

LeanDojo~\cite{yang2024leandojo} boosts theorem proving by using retrieval-augmented methods to choose relevant premises from extensive mathematical libraries, improving automation and theorem generalization.
RAG-for-math-QA~\cite{levonian2023retrieval} improves math question-answering by integrating a high-quality math textbook with RAG, enhancing LLM-generated responses for middle-school algebra and geometry.

\section{Benchmark}
\label{sec:benchmark}

Given the increasing research interests and applications of RAG, there have also been several benchmarks assessing RAG from certain aspects.

Chen et al.~\cite{RAG_Benchmark} proposed an RAG benchmark that evaluates across four dimensions: 
(1) Noise Robustness, testing if LLMs can extract necessary information from noisy documents;
(2) Negative Rejection, assessing if LLMs can reject to respond when retrieved content is insufficient;
(3) Information Integration,checking if LLMs can acquire knowledge and respond by integrating multiple retrieved contents;
(4) Counterfactual Robustness, determining if LLMs can identify counterfactual errors in retrieved content.

Three other benchmarks, RAGAS~\cite{RAGAS}, ARES~\cite{ARES}, and TruLens~\cite{TurLens}, evaluate three aspects using a separate evaluator LLM: (1) Faithfulness, assessing factual accuracy based on retrieved content; (2) Answer Relevance, determining if results address the queries; (3) Context Relevance, evaluating the relevance of retrieved content and its conciseness.

CRUD-RAG~\cite{CRUD-RAG} divides RAG tasks into four types: Create, Read, Update, and Delete, assessing them through text continuation, question answering, hallucination correction, and open-domain multi-document summary.
MIRAGE~\cite{MIRAGE} assesses RAG in the medical domain, focusing on the performance of medical question-answering systems.
KILT~\cite{KILT} aligns Wikipedia snapshots to verify information accuracy, using BLEU scores to pinpoint relevant texts and filtering to uphold quality, thus providing diverse retrieval systems for evidence-backed predictions or citations.

\section{Discussion}
\label{sec:discussion}

\subsection{\textbf{Limitations}}

Despite the widespread adoption of RAG, 
it suffers from several limitations by nature.

\subsubsection{\textbf{Noises in Retrieval Results}}

Information retrieval is inherently flawed due to information loss in item representations and ANN search.
The inevitable noise, manifesting as irrelevant content or misleading information, can create failure points in RAG systems~\cite{DBLP:journals/corr/abs-2401-05856}.
However, although improving retrieval accuracy seems intuitive for RAG effectiveness, recent research surprisingly finds that noisy retrieval results might enhance generation quality~\cite{DBLP:journals/corr/abs-2401-14887}.
A possible explanation is that diverse retrieval outcomes could contribute to prompt construction~\cite{qiu2022evaluating}.
Thus, the impact of retrieval noise remains unclear, leading to confusion about metric selection and retriever-generator interaction in practical uses.

\subsubsection{\textbf{Extra Overhead}}

While retrieval can reduce generation costs in certain cases~\cite{Atlas,MemTransformer2022,REST}, it incurs non-negligible overhead in most cases.
%
In other words, the retrieval and interaction processes increase latency inevitably. 
This is amplified when RAG is combined with complex enhancement methods, such as recursive retrieval~\cite{Query_Expansion_by_Prompting_LLMs} and iterative RAG~\cite{DBLP:conf/emnlp/ZhangCZKLZMLC23}.
Furthermore, as the scale of retrieval sources expands, the storage and access complexity will also increase~\cite{EA}. 
Such overhead hampers the practicality of RAG in real-time services that are sensitive to latency.

\subsubsection{\textbf{The Gap between Retrievers and Generators}}

Since the objectives of retrievers and generators may not align, and their latent spaces might differ, designing their interaction requires meticulous design and optimization.
Current approaches either disentangle retrieval and generation or integrate them at an intermediate stage.
While the former is more modular, the latter could benefit from joint training but hamper 
generality.
Selecting a cost-effective interaction method to bridge the gap poses a challenge and necessities deliberation in practice.

\subsubsection{\textbf{Increased System Complexity}}
 
The introduction of retrieval unavoidably increases the system complexity and the number of hyper-parameters to tune.
For instance, a recent study found that using top-$k$ rather than a single retrieval improves attribution but harms fluency in query-based RAG~\cite{DBLP:journals/corr/abs-2302-05578},
while other aspects such as metric selection are still under explored.
Thus, it requires more expertise to tune the generation service when RAG is involved.

\subsubsection{\textbf{Lengthy Context}}
One of the primary shortcomings of RAG, in particular the query-based RAG, is that it lengthens the context tremendously, making it infeasible for generators with limited context length.
In addition, the lengthened context also slows down the generation process generally.
The research advancements in prompt compression~\cite{LLMLingua} and long-context support~\cite{DBLP:journals/corr/abs-2308-16137} have partially mitigated these challenges, albeit with a slight trade-off in accuracy or costs.

\subsection{\textbf{Potential Future Directions}}

Lastly, we wish to outline several potential directions for future RAG research and applications.

\subsubsection{\textbf{Novel Design of Augmentation Methodologies}}

Existing research has explored various interaction patterns between retrievers and generators.
However, due to distinct objectives in these two components, the practical augmentation process has a significant impact on the final generation results.
Investigation of more advanced foundations for augmentation holds promise for fully unleashing the potential of RAG.

\subsubsection{\textbf{Flexible RAG Pipelines}}

RAG systems are progressively embracing flexible pipelines, such as recursive, adaptive, and iterative RAG.
With precise tuning and meticulous engineering, the unique blend of retrieval sources, retrievers, generators, and RAG subsystems promises to tackle complex tasks and boost overall performance.
We eagerly anticipate pioneering exploration that will drive the evolution of even more innovative RAG systems.

\subsubsection{\textbf{Broader Applications}}

RAG is a general technique applied in various applications.
However, some generative tasks have not yet explored RAG, and in many domains, RAG is applied naively without considering the domain's unique characteristics.
We believe designing domain-specific RAG techniques will significantly benefit broader applications.

\subsubsection{\textbf{Efficient Deployment and Processing}}

There exist several deployment solutions for query-based RAG with LLMs, such as LangChain~\cite{langchain}, LLAMA-Index~\cite{LlamaIndex}, and PipeRAG~\cite{jiang2024piperag}.
However, for other RAG foundations and/or generation tasks, there lacks a plug-and-play solution. 
Besides, due to retrieval overhead and increasing complexities in retrievers and generators, achieving efficient RAG is still challenging and necessitates further system-level optimizations.

\subsubsection{\textbf{Incorporating Long-tail and Real-time Knowledge}}

While a key motivation of RAG is to harness real-time and long-tail knowledge, few studies have explored the pipeline for knowledge updating and expansion.
Many existing works use merely the generators' training data as retrieval sources, neglecting the dynamic and flexible information that retrieval could offer.
As a consequence, there is a growing research on designing RAG systems with continuously updated knowledge and flexible sources.
We also expect RAG to step further, adapting to personalized information in today's web service.

\subsubsection{\textbf{Combined with Other Techniques}}

RAG is orthogonal to other techniques that also aim to improve AIGC effectiveness, such as fine-tuning, reinforcement learning, chain-of-thought, and agent-based generation.
The combining of these methods~\cite{meduri2024efficient} is still in its early stages, calling for further research to fully exploit their potential through novel algorithm designs.
It is worthy to note that a recent notion appears ``long-context models like Gemini 1.5 will replace RAG''.
Nevertheless, this assertion overlooks RAG's flexibility in managing dynamic information, encompassing both up-to-date and long-tail knowledge~\cite{longcontextkillRAG}.
We expect RAG to benefit from long context generation, rather than being replaced by it.

\section{Conclusion}
\label{sec:conclusion}

In this paper, we conducted a thorough and comprehensive survey on RAG within the context of AIGC, with a particular focus on augmentation foundations, enhancements, and applications.
We first systematically organized and summarized the foundation paradigms in RAG, providing insights into the interaction between retrievers and generators.
%
Then, we reviewed the enhancements that further improve the effectiveness of RAG, including the enhancements on each component or the entire pipeline.
To facilitate researchers across diverse domains, we showcased practical applications of RAG in a range of modalities and tasks.
Finally, we also presented existing benchmarks for RAG, discussed current limitations of RAG, and shed light on promising future directions.
\bibliography{reference}
\bibliographystyle{IEEEtran}

\end{document}